\documentclass[lettersize,journal]{IEEEtran}
\usepackage{amsmath,amsfonts}
\usepackage{algorithmic}
\usepackage{algorithm}
\usepackage{array}
\usepackage[caption=false,font=normalsize,labelfont=sf,textfont=sf]{subfig}
\usepackage{textcomp}
\usepackage{stfloats}
\usepackage{url}
\usepackage{verbatim}
\usepackage{graphicx}
\usepackage{booktabs}
\usepackage{multicol}
\usepackage{multirow}
\usepackage{colortbl}
\usepackage[table]{xcolor}
\definecolor{Gray}{gray}{0.9} 
\usepackage{cite}
\usepackage{amssymb}
\begin{document}

\title{Frequency-Domain Decomposition and Recomposition for Robust Audio-Visual Segmentation}

\author{Kai Peng\#, Yunzhe Shen\# , Miao Zhang$^{*}$, \IEEEmembership{Member, IEEE}, Jia Wang, Leiye Liu, Wei Ji, Jingjing Li,   \\Yongri Piao$^{*}$, ~\IEEEmembership{Member, IEEE} and Huchuan Lu, \IEEEmembership{Fellow, IEEE}
\thanks{This paper was produced by the IEEE Publication Technology Group. They are in Piscataway, NJ. Manuscript received April 19, 2021; revised August 16, 2021. This work was supported in part by the National Natural Science Foundation of China under Grant 62372080, Grant 62376050, and in part by the Natural Science Foundation of Liaoning Province under Grant 2024-MSBA-24. (\# is equal contribution, * is corresponding author)}
\thanks{Kai Peng, Yunzhe Shen,  Miao Zhang, Jia Wang, Leiye Liu,  Yongri Piao, and Huchuan Lu are with Dalian University of Technology, Dalian 116024, China (e-mail: happypk@mail.dlut.edu.cn; 1079006\allowbreak460\allowbreak@mail.dlut.edu.cn; miaozhang\allowbreak@dlut.edu.cn; Jiawang0704\allowbreak@outlook.com; leiyeliu\allowbreak@mail.dlut.edu.cn;  yrpiao\allowbreak@dlut.edu.cn; lhchuan\allowbreak@dlut.edu.cn).}
\thanks{Wei Ji is with the School of Medicine, Yale University, New Haven,
CT 06520 USA (e-mail: wei.ji@yale.edu).}
\thanks{Jingjing Li is with the School of Computer Science,
Carnegie Mellon University, Pittsburgh, PA 15213, USA
(e-mail: jingjingli.cmu@gmail.com).}
}
\markboth{IEEE Transactions on Image Processing}
{Peng \MakeLowercase{\textit{et al.}}:
Frequency-Domain Decomposition and Recomposition
for Robust Audio-Visual Segmentation}


\maketitle

\begin{abstract}
Audio-visual segmentation (AVS) plays a critical role in multimodal machine learning by effectively integrating audio and visual cues to precisely segment objects or regions within visual scenes. Recent AVS methods have demonstrated significant improvements. However, they overlook the inherent frequency-domain contradictions between audio and visual modalities--the substantial noise often present in audio high-frequency signals vs. the structurally rich details in visual high-frequency signals. Ignoring these differences can result in suboptimal performance. In this paper, we rethink the AVS task from a deeper perspective by reformulating AVS task as a frequency-domain decomposition and recomposition problem. To this end, we introduce a novel Frequency-Aware Audio-Visual Segmentation (FAVS) framework consisting of two key modules: Frequency-Domain Enhanced Decomposer (FDED) module and Synergistic Cross-Modal Consistency (SCMC) module. FDED module employs a residual-based iterative frequency decomposition to discriminate modality-specific semantics and structural features, and SCMC module leverages a mixture-of-experts architecture to reinforce semantic consistency and modality-specific feature preservation through dynamic expert routing. Extensive experiments demonstrate that our FAVS framework achieves state-of-the-art performance on three benchmark datasets, and abundant qualitative visualizations further verify the effectiveness of the proposed FDED and SCMC modules. The project repository is available at \url{https://github.com/happylife-pk/FAVS}.
\end{abstract}


\begin{figure}[t]
  \centering
  \includegraphics[width=\linewidth]{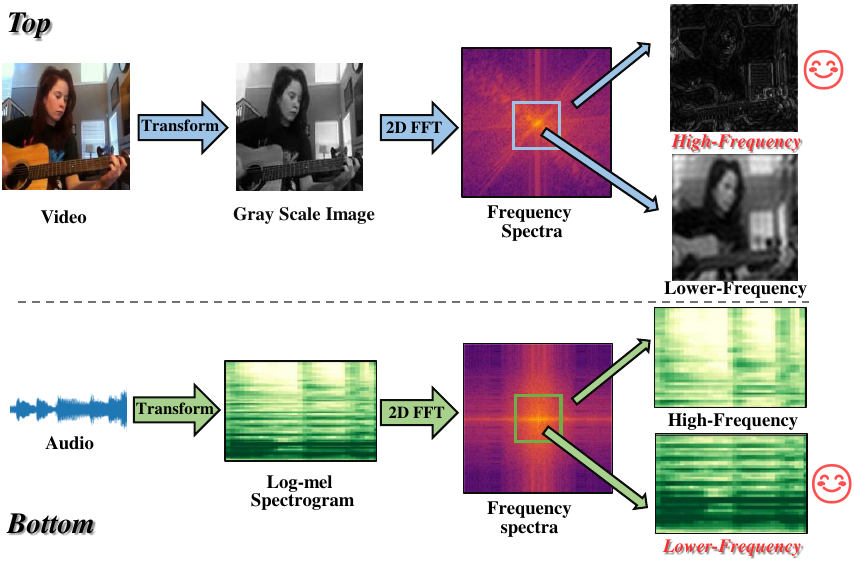}
  \caption{Illustration of frequency-domain contradictions in AVS. Frequency spectra depict signal strength, with inner regions representing lower frequencies and outer regions corresponding to higher frequencies. In video (Top), high frequencies (outer regions) preserve critical structural details (e.g., object edges) for precise segmentation. Conversely, in audio (Bottom), high frequencies (outer regions)
often contain substantial noise,
while lower frequencies contain rich semantic information.}
  \label{moti}
\end{figure}

\section{Introduction}

\IEEEPARstart{T}{he} audio-visual segmentation (AVS) task plays a crucial role in multimodal machine learning by integrating auditory and visual cues to accurately segment objects in visual scenes~\cite{zhou2022audio}. Recent advancements in AVS have been driven by query-based architectures~\cite{yang2024cooperation}~\cite{ma2024stepping}~\cite{sun2024unveiling}, which leverage cross-modal fusion and attention mechanisms to effectively combine audio and video features, enhancing segmentation performance beyond early AVS methods built upon traditional unimodal architectures such as standard Transformers~\cite{ling2024transavs}~\cite{hao2024improving}, U-Net~\cite{mao2025contrastive}, and DeepLab~\cite{li2023catr}~\cite{shi2024cross}.

However, despite these advancements, a less explored yet critical challenge is overlooked: There are inherent frequency-domain contradictions between audio and video signals, as illustrated in Fig.~\ref{moti}. In audio signals, high-frequency components often contain substantial noise, which can degrade segmentation accuracy if not properly suppressed. In contrast, high-frequency components in video capture essential structural details, such as object edges, which are critical for precise segmentation. Fusing the two modalities without accounting for their frequency-domain differences can weaken important features in one modality, leading to dislocation of sound sources or blurred object boundaries. 

In rethinking the significance of the frequency domain in the audio-visual segmentation process, we argue that it indeed reflects important information for both modalities~\cite{zhang2023fsanet}. Recent methods have aimed to establish a correlation between audio-visual modalities by leveraging the spatio-temporal consistency. For example, AVSAC~\cite{chen2025bootstrapping} designs a loss function to address spatio-temporal alignment, and CATR~\cite{li2023catr} introduces a fusion module that combines decoupled audio-visual features across spatio-temporal dimensions. These methods have achieved excellent performance. However, they do not consider the  differences in the modality frequency domain. This might explain why existing methods perform less effectively in complex scenarios, such as multi-source environments or overlapping sound sources. DDESeg~\cite{liu2025dynamic} observes that overlapping audio signals fail to capture distinct semantic characteristics of each sound, causing unreliable audio-visual alignment. It addresses this by identifying distinct semantic representations through enriching the incomplete information of individual sound sources. But DDESeg does not explicitly explore the frequency-domain issues between audio and visual modalities. Therefore, a critical research challenge remains: how do we effectively resolve the inherent frequency-domain contradiction?

To address this issue, we formulate the AVS task as a frequency decomposition and recomposition problem. In the decomposition phase, it is essential to precisely isolate modality-specific semantic and structural information across distinct frequency bands, while in the recomposition phase, the emphasis shifts to synergistically integrating these bands to obtain coherent and complementary cross-modal feature representations. To this end, we propose a novel Frequency-Aware Audio-Visual Segmentation (FAVS) framework. FAVS comprises two key modules: a Frequency-Domain Enhanced Decomposer (FDED) module and a Synergistic Cross-Modal Consistency (SCMC) module. 

FDED module employs a residual iteration strategy in frequency decomposition to prevent information loss, thereby effectively capturing the modality semantic information in the frequency domain. Specifically, we perform Fourier-based decomposition on both signals (original bands) to four frequency bands (high, middle, low-frequency bands, and the residual band), since learning on the frequency spectrum enables the model to manipulate all frequency bands and capture structural semantic details. This approach extracts informative audio features from the low-to-middle frequencies (e.g., classification), while preserving essential visual features from high-frequencies (e.g., object edges and textures). Through this process, the semantic and structural information of the two modalities is discriminatively separated and fully exploited in the frequency domain.

To enforce consistent representations between audio and visual modalities in the frequency domain, we propose the Synergistic Cross-Modal Consistency (SCMC) module. Specifically, SCMC module excavates cross-modal correlations among decomposed frequency components by employing the mixture-of-experts (MoE) on cross-modal attention, thereby capturing consistent semantic information across modalities. Within SCMC module, each attention expert learns to specialize in capturing useful features in distinct frequency bands. The routing mechanism dynamically integrates experts for ensuring optimal cross-modal synergy strategies, leveraging complementary features from one modality to preserve crucial features that may be misleadingly suppressed in the other modality during decomposition. SCMC module not only reinforces consistent representations but also preserves special yet critical modality-specific features.

In summary, we are the first to explicitly integrate frequency-domain decomposition and recomposition into the AVS task, effectively addressing inherent frequency-domain contradictions between audio and visual modalities through our proposed FAVS framework, and achieving state-of-the-art (SOTA) performance across three benchmark AVS datasets (S4, MS3 and AVSS). Meanwhile, extensive qualitative visualizations further verify the effectiveness of our proposed FDED and SCMC modules.

\section{Related Work}

\subsection{Audio-Visual Segmentation}

Audio-visual learning has progressed significantly in recent
years~\cite{zhu2025does}, and Audio-Visual Segmentation (AVS) is an extended multimodal task that aims to generate fine-grained pixel-level segmentation masks for sounding objects in video frames. The fusion-oriented frameworks integrate audio and visual encoder outputs before decoding, enabling pixel-wise predictions; for instance, TPAVI~\cite{zhou2022audio} employs attention-driven aggregation across encoder stages, followed by a Feature Pyramid Network-like decoder~\cite{chen2017deeplab} for multi-scale refinement. Moreover, AVSAC~\cite{chen2025bootstrapping} mitigates the disparity between modalities stemming from one-sided and inadequate incorporation of auditory signals. It proposes a bidirectional audio-visual decoder equipped with integrated bridges to amplify audio features, addressing the modality imbalance generally exists in multimodal fields~\cite{han2022trusted,liu2023multi,zhang2022c}. Additionally, learning-based methods have been widely used in dense prediction tasks~\cite{piao2022semi,ren2022adaptive,zhang2022exploring}, and more recent Transformer-driven approaches~\cite{wang2024ref,chen2024cpm,liu2023audio} adopt a mask-query paradigm inspired by Mask2Former, where audio-informed queries traverse decoder layers to identify sonic objects. Notably, COMBO~\cite{yang2024cooperation} augments trainable queries with auditory embeddings and conducts bidirectional audio-visual merging to refine query representations. However, the overlapping nature of audio creates inherent ambiguity, complicating the relation between auditory cues and visual regions. Recent studies~\cite{li2024qdformer}~\cite{li2023catr} mitigate this challenge through sound separation techniques that decompose mixed audio streams, or decouple audio-visual features across spatio-temporal dimensions. 

Additionally, recent approaches explore efficient adaptation
and alternative query designs. ALOHA~\cite{zhou2025aloha} adapts
local spatio-temporal context, METTLE~\cite{zhou2025mettle} learns
compact meta-tokens for memory-efficient audio-visual adaptation,
and VCT~\cite{huang2025revisiting} constructs vision-derived queries for
sounding-object segmentation. Crab~\cite{du2025crab} extends unified
multimodal understanding to pixel-level tasks through explicit
cooperation.
However, these methods lack exploration in terms of the frequency
domain, which may lead to performance bottlenecks under noisy
conditions or multi-source scenes.

\subsection{Mixture of Experts}

Mixture of Experts (MoE) architectures have emerged as a powerful paradigm for scaling large models while maintaining efficiency through sparse activation and specialized sub-networks~\cite{riquelme2021scaling,du2022glam,wang2025m2restore}. In video processing, MoE has been applied to enhance zero-shot video question answering via temporal-guided routing~\cite{sun2020drcnn} and to improve video deblurring through attentive large kernel networks integrated with dynamic expert mechanisms~\cite{meng2025br,abbas2020biased}. In multimodal learning, MoE facilitates adaptive integration of diverse inputs, such as vision and language, by routing tokens to domain-specific experts, as demonstrated in Uni-MoE~\cite{li2025uni}, which scales unified multimodal large language models (MLLMs) across text, image, and audio modalities via expert mixtures for enhanced generalization. Extending to the AV-related task, AVMoE~\cite{cheng2024mixtures} explores parameter-efficient transfer learning by injecting MoE adapters into pretrained audio-visual models, outperforming baselines in AVS, event localization, and question answering via modality-robust fusion. Inspired by MoE, we design SCMC module, each expert in the module is trained to specialize in processing distinct frequency bands enriched with audio-visual features, allowing for consistent learning of frequency-specific cues. 
\begin{figure*}[t]
  \centering
  \includegraphics[width=\linewidth]{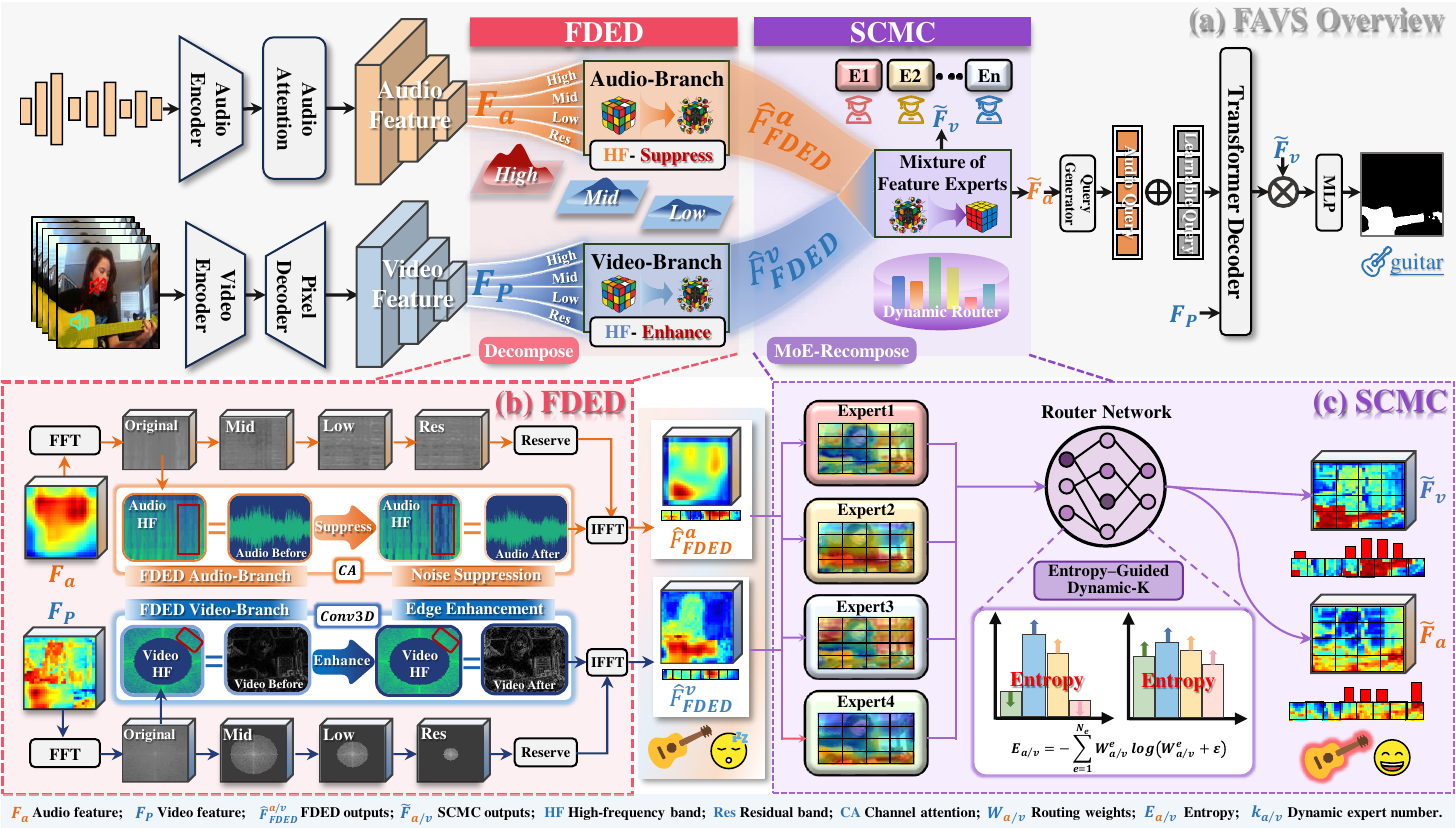}
  \caption{Overview of the proposed FAVS framework and its key components: (a) \textbf{Overview Architecture}. The framework processes video and audio inputs through their corresponding encoders, and progressively refines visual and audio features via three stages of FDED and SCMC modules. The refined audio features are further used to generate audio queries, which are fused with learnable queries in the transformer decoder to predict the final segmentation masks. (b) \textbf{FDED Module}. This module transforms modality features into the frequency domain, performs residual radial decomposition to separate high-, mid-, low-, and residual-frequency bands, and applies modality-specific heterogeneous operations on the high-frequency bands, i.e., suppressing noisy audio high-frequency components while enhancing discriminative visual high-frequency structures, followed by inverse transformation and weighted reconstruction. (c) \textbf{SCMC Module}. This module employs a mixture-of-experts architecture to model complementary cross-modal patterns across the reconstructed frequency bands. An entropy-guided dynamic routing mechanism adaptively determines the number of activated experts and aggregates their outputs to restore suppressed features and enhance cross-modal consistency.}
  \label{DAMF}
\end{figure*}

\section{Methods}

\subsection{Architecture Overview}

We first describe the overall architecture of the FAVS shown in Fig.~\ref{DAMF}, which leverages the transformer architecture~\cite{cheng2022masked}. We first feed the video frames into the visual backbone SwinB~\cite{liu2021swin} to extract multi-scale visual features. For each input video $V \in \mathbb{R}^{T\times3\times H \times W}$ with $T$ frames, the visual backbone extracts multi-scale image features $F_v=[V_m]^{3}_{m=1}$. And then, we employ the pixel decoder~\cite{zhudeformable} to construct discriminative, high-resolution per-pixel embeddings $F_p=[P_i]_{i=1}^{3}$, where $P_i \in \mathbb{R}^{T\times C\times H_i \times W_i}$, $(H_i, W_i)=(H, W)/2^{i+1}$, and $C$ are the height, width, and channel numbers of the visual feature from the $i$-th stage of the output. Meanwhile, the original audio clip is first converted to a mono output, and a short-time Fourier transform is performed to yield a mel spectrum $A\in \mathbb{R}^{T\times96\times64}$. Furthermore, the audio backbone VGGish~\cite{hershey2017cnn} utilizes $A$ as input to generate audio features $F_a \in \mathbb{R}^{T\times C\times H_3\times W_3}$.

The Frequency-Aware Audio-Visual Segmentation (FAVS) framework consists of three progressive stages, utilizing $F_p$ and $F_a$ as inputs, with each stage incorporating both Frequency-Domain Enhanced Decomposer (FDED) module and Synergistic Cross-Modal Consistency (SCMC) module. The workflow is mathematically expressed as:
\begin{equation}
\begin{gathered}
\hat F^{i}_{v},\hat F^{i}_{a} =
\begin{cases}
\Phi_{F}(P_{i}, F_{a}), & i=1,\\
\Phi_{F}(Conv([\tilde F^{i-1}_{v}, P_{i}]), \tilde F^{i-1}_{a}), & i>1, 
\end{cases}  \\
\tilde F^{i}_{v},\tilde F^{i}_{a} = \Phi_{S}(\hat F^{i}_{v}, \hat F^{i}_{a}).
\end{gathered}
\end{equation}
where $\hat F^{i}_{v}$ and $\hat F^{i}_{a}$ denote the outputs of the FDED module at stage $i$, while $\tilde F^{i}_{v}$ and $\tilde F^{i}_{a}$ denote the outputs of the SCMC module. $\Phi_F$ and $\Phi_S$ correspond to the operations of the FDED and SCMC modules, respectively.

After these two modules, we derive the object queries added with learnable embeddings from $\tilde F^{3}_{a}$ through global average pooling followed by a channel-wise $MLP$. Finally, the outputs of these two branches are fused within a transformer decoder, and the final object masks are predicted by a prediction head. 

\subsection{Frequency-Domain Enhanced Decomposer}

FDED module decomposes modality-specific semantic and structural information by transforming audio and visual features into the frequency domain. The motivation of the module is to suppress noise in audio while enhancing edge structure information in visuals, preserving important features in both modalities.

Specifically, the input features $F_{input} \in \{F_p, F_a\}$ from both modalities are initially processed through convolutional layers, incorporating depthwise separable convolutions $DWC$ and grouped 1$\times$1 convolutions $GroupConv$. These convolutional operations are specifically designed to preprocess the feature maps in the spatial domain, enhancing both their semantic and spatial representations. Following this, a two-dimensional Fast Fourier Transform (2D FFT) $\mathcal{F}$ is applied to convert the processed feature maps $F_{pre}$ into the frequency domain $\mathcal{F}_{pre}$, shown as:
\begin{equation}
\begin{gathered}
F_{pre} = GroupConv(DWC(F_{input})), \\
\mathcal{F}_{pre} = \mathcal{F}(F_{pre}).
\end{gathered}
\end{equation}

Inspired by the intrinsic mode function (IMF)~\cite{barbosh2020empirical}, we decompose the frequency spectrum into distinct frequency bands using a residual mechanism to prevent information loss. Specifically, starting from the original frequency spectrum $\mathcal{F}_{original}$, we select suitable frequency thresholds $\tau_{i}$ and extract frequency bands $\mathcal{F}_{band}^{i}$ from the current remaining frequency spectrum $\mathcal{F}_{rem}^{i}$. Formally, this residual iteration can be represented as:
\begin{equation}
\begin{gathered}
\mathcal{F}_{rem}^{0} = \mathcal{F}_{original} = \mathcal{F}_{pre}, \\
\mathcal{F}_{rem}^{i} = \mathcal{F}_{rem}^{i-1} - \mathcal{F}_{band}^{i},  \quad i = 1, 2, \ldots, N,\\
\mathcal{F}_{band}^{i}(\omega) =
\begin{cases}
\mathcal{F}_{rem}^{i-1}(\omega), & \tau_i < |\omega| \leq \tau_{i-1}, \\
0, & \text{otherwise},
\end{cases} \\
\mathcal{F}_{res} = \mathcal{F}_{pre} - \sum_{i=1}^{N}\mathcal{F}_{band}^{i}.
\end{gathered}
\end{equation}
Here, $|\omega|$ represents the normalized radial frequency magnitude. It is defined as the radial distance from the origin in the frequency spectrum, normalized by the maximum radius, i.e., $|\omega| = \sqrt{u^{2} + v^{2}} / r_{\max} \in [0,1]$, where $(u, v)$ are the frequency-plane coordinates and $r_{\max}$ denotes the maximum radial distance in the spectrum. This formulation implies that frequencies near the center represent lower frequencies, while those farther away correspond to higher frequencies~\cite{glisson2025hsda}. We set $\tau_0 = 1$ and $\tau_0 > \tau_1 > \cdots > \tau_N > 0$. Unless otherwise stated, the default three-band decomposition adopts $(\tau_1,\tau_2)=(0.7,0.3)$, where the extracted frequency bands explicitly correspond to the high-, mid-, and low-frequency bands, i.e., $\mathcal{F}_{h}=\mathcal{F}_{band}^{1}$, $\mathcal{F}_{m}=\mathcal{F}_{band}^{2}$, and $\mathcal{F}_{l}=\mathcal{F}_{band}^{3}$. Other threshold settings are evaluated in the ablation study. $\mathcal{F}_{res}$ is the final residual frequency band.

Then we apply a 3D convolution ($Conv3D$) to the video modality’s high-frequency band $\mathcal{F}_{h}^{v}$ to strengthen these details, while employing a channel attention mechanism $CA$ on the audio modality's high-frequency band $\mathcal{F}_{h}^{a}$ to suppress disruptive noise components. 3D convolutions have been demonstrated to effectively enhance edge-related high-frequency structures in videos~\cite{diba2023spatio}. The mid-frequency and low-frequency bands for both modalities are preserved without further manipulation. These modality-specific enhancements are formally expressed as:
\begin{equation}
\begin{gathered}
\hat{\mathcal{F}}_{h}^{v} = Conv3D(\mathcal{F}_{h}^{v}) + \mathcal{F}_{h}^{v}, \\
\hat{\mathcal{F}}_{h}^{a} = CA(\mathcal{F}_{h}^{a}) \odot \mathcal{F}_{h}^{a},
\end{gathered}
\end{equation}
where $\hat{\mathcal{F}}_{h}^{v}$ and $\hat{\mathcal{F}}_{h}^{a}$ represent the enhanced high-frequency bands. Since high-frequency audio can also contain useful
semantic cues, this discriminative channel-wise reweighting aims
to suppress irrelevant responses while retaining informative
components for subsequent recomposition. 

Finally, the enhanced high-frequency bands $\hat{\mathcal{F}}_{h}^{a/v}$ and others $\mathcal{{F}}_{band}^{a/v}$ ($band \in{mid, low, residual}$) are individually transformed back into the spatial domain $\hat{F}_{Band}^{a/v}$ ($Band\in\{h,m,l,res\}$) using the inverse 2D FFT operation $\mathcal{F}^{-1}$:
\begin{equation}
\begin{gathered}
\hat{F}_{Band}^{a/v} =
\begin{cases}
\mathcal{F}^{-1}(\hat{\mathcal{F}}_{h}^{a/v}), & band = h,\\
\mathcal{F}^{-1}(\mathcal{F}_{band}^{a/v}), & band \in \{m, l, res\}.
\end{cases}
\end{gathered}
\end{equation}

To reconstruct the final feature maps $\hat{F}_{FDED}^{a/v}$ from these frequency bands, we apply modality-specific weighted summations $w_{Band}^{a/v}$ to ensure effective preservation of critical frequency information, shown as:
\begin{equation}
\begin{aligned}
\hat{F}_{FDED}^{v} &= \sum_{Band \in \{h,m,l,res\}} w_{Band}^{v}\hat{F}_{Band}^{v}, \\
\hat{F}_{FDED}^{a} &= \sum_{Band \in \{h,m,l,res\}} w_{Band}^{a}\hat{F}_{Band}^{a},
\end{aligned}
\end{equation}
where $\hat{F}_{FDED}^{v}$ and $\hat{F}_{FDED}^{a}$ denote the reconstructed spatial-domain video and audio features. The weights $w_{Band}^{a/v}$ are learned separately
for the audio and visual modalities by applying softmax to trainable
coefficients, ensuring nonnegative weights that sum to one within
each modality. These coefficients are updated jointly with the
network through backpropagation of the overall training loss.

\begin{table*}[t!]
\footnotesize
\caption{Comparison With State-of-the-Art Methods on the S4 and MS3 Datasets. The best results are highlighted in \textbf{bold}, and the second-best are \underline{underlined}. $\mathcal{J}\&\mathcal{F}_m$ means the average value of $\mathcal{J}$ and $\mathcal{F}_m$. \textsuperscript{*} Indicates results based on the evaluation criteria introduced by DDESeg~\cite{liu2025dynamic}. Our corresponding result (FAVS\textsuperscript{*}) follows the same criteria for a fair comparison. --: unreported. \textsuperscript{$\dagger$} METTLE uses $192\times192$ visual inputs.}
\label{AVSexp}
\centering
\setlength{\tabcolsep}{8pt}
\renewcommand{\arraystretch}{1.1}
\begin{tabular}{l c c ccc ccc}
\toprule
\multirow{2}{*}{Method} &
\multirow{2}{*}{Reference} &
\multirow{2}{*}{Backbone} &
\multicolumn{3}{c}{S4} &
\multicolumn{3}{c}{MS3} \\
\cmidrule(lr){4-6}\cmidrule(lr){7-9}
& & &
$\mathcal{J}$ &
$\mathcal{F}_m$ &
$\mathcal{J}\&\mathcal{F}_m$ &
$\mathcal{J}$ &
$\mathcal{F}_m$ &
$\mathcal{J}\&\mathcal{F}_m$ \\
\midrule
\multicolumn{9}{l}{\textbf{\textit{Results on 224$\times$224 Image Size}}}\\
\midrule
TPAVI & ECCV'22 & PVT-v2 & 78.7 & 87.9 & 83.3 & 54.0 & 64.5 & 59.3 \\
AQFormer & IJCAI'23 & PVT-v2 & 81.6 & 89.4 & 85.5 & 61.1 & 72.1 & 66.6 \\
CATR & ACMMM'23 & PVT-v2 & 81.4 & 89.6 & 85.5 & 57.8 & 70.8 & 64.3 \\
AVSC & ACMMM'23 & Swin-Base & 81.3 & 88.6 & 84.9 & 59.5 & 65.8 & 62.7 \\
ECMVAE & ICCV'23 & PVT-v2 & 81.7 & 90.1 & 85.9 & 57.8 & 70.8 & 64.3 \\
BAVS & TMM'24 & Swin-Base & 82.7 & 89.8 & 86.3 & 59.6 & 65.9 & 62.8 \\
AVSBG & AAAI'24 & PVT-v2 & 81.7 & 90.4 & 86.1 & 55.1 & 66.8 & 61.0 \\
GAVS & AAAI'24 & ViT-Base & 80.1 & 90.2 & 85.2 & 63.7 & 77.4 & 70.6 \\
AVSegformer & AAAI'24 & PVT-v2 & 82.1 & 89.9 & 86.0 & 58.4 & 69.3 & 63.8 \\
UFE & CVPR'24 & PVT-v2 & 83.2 & 90.4 & 86.8 & 62.0 & 70.9 & 66.5 \\
COMBO & CVPR'24 & PVT-v2 & 84.7 & 91.9 & 88.3 & 59.2 & 71.2 & 65.2 \\
AVSStone & ECCV'24 & Swin-Base & 83.2 & 91.3 & 87.3 & 67.3 & 77.6 & 72.4 \\
SelM & ACMMM'24 & PVT-v2 & 83.5 & 91.2 & 87.4 & 60.3 & 71.3 & 65.8 \\
AVSAC & TCSVT'25 & PVT-v2 & 84.5 & 91.6 & 88.1 & 64.2 & 76.6 & 70.4 \\
DDESeg\textsuperscript{*} & CVPR'25 & PVT-v2 &
\underline{90.3} & 92.2 & \underline{91.3} &
69.7 & \underline{79.6} & 74.7 \\
DiffusionAVS & TIP'25 & PVT-v2 & 81.5 & 90.3 & 85.9 & 59.6 & 71.2 & 65.4 \\

ALOHA &
TOMM'25 &
CAFormer-B36 &
82.1 &
90.1 &
86.1 &
58.6 &
70.6 &
64.6 \\

METTLE\textsuperscript{$\dagger$} &
TPAMI'25 &
Swin-V2-L &
80.7 &
-- &
-- &
55.1 &
-- &
-- \\

VCT &
CVPR'25 &
Swin-Base &
84.7 &
\underline{92.3} &
88.5 &
67.5 &
79.3 &
73.4 \\

Crab &
CVPR'25 &
ViT-L/14 &
73.25 &
86.81 &
80.03 &
58.21 &
66.24 &
62.23 \\

\rowcolor{red!4}
\textbf{FAVS (Ours)} & -- & Swin-Base &
85.6 & 92.2 & 88.9 &
\underline{71.1} & 78.7 & \underline{74.9} \\
\rowcolor{red!12}
\textbf{FAVS (Ours)}\textsuperscript{*} & -- & Swin-Base &
\textbf{92.7} & \textbf{94.0} & \textbf{93.4} &
\textbf{77.0} & \textbf{86.1} & \textbf{81.6} \\
\midrule
\multicolumn{9}{l}{\textbf{\textit{Results on 384$\times$384 Image Size}}}\\
\midrule
TeSo & ECCV'24 & Swin-Base & 83.3 & 93.3 & 88.3 & 66.0 & 80.1 & 73.1 \\
AVSBias & ACMMM'24 & Swin-Base & 83.3 & 93.0 & 88.2 & 67.2 & 80.8 & 74.0 \\
CQFormer & TIP'25 & PVT-v2 & 83.6 & 91.2 & 87.4 & 61.0 & 72.7 & 66.9 \\
\rowcolor{blue!6}
\textbf{FAVS (Ours)} & -- & Swin-Base &
\textbf{85.7} & \textbf{93.3} & \textbf{89.5} &
\textbf{74.4} & \textbf{81.7} & \textbf{78.1} \\
\bottomrule
\end{tabular}
\end{table*}

\subsection{Synergistic Cross-Modal Consistency}

 Synergistic Cross-Modal Consistency (SCMC) module applies mixture-of-experts (MoE) architecture to cross-modal attention, enabling each expert to specialize in capturing complementary features from distinct frequency bands. To this end, it reinforces semantic consistency across modalities while preserving critical modality-specific features. Specifically, SCMC module employs $N_e=4$ attention experts, with a dynamic routing mechanism that adaptively selects and weights experts.  Here, $N_e$ denotes the pool size, while the number activated for
each input is determined dynamically.

For each expert, given the reconstructed visual features $\hat{F}_{FDED}^{v}$ and audio features $\hat{F}_{FDED}^{a}$, we first apply a Spatial-Temporal-Channel feature enhancer ($STC$), which facilitates specialization by adaptively emphasizing frequency-specific auditory cues and their visually correlated patterns. This is achieved through joint modeling of spatial structures, temporal dynamics, and channel dependencies. Operationally, it can be defined as:
\begin{equation}
\begin{gathered}
F_{ST}^{a/v} = DWConv(\hat{F}_{FDED}^{a/v}),\\
F_{STC}^{a/v} = LN(F_{CAR}^{a/v}),
\end{gathered}
\end{equation}
where $DWConv$~\cite{chollet2017xception} denotes the depthwise 3D convolution used to extract more semantically rich and computationally efficient spatio-temporal representations $F_{ST}^{a/v}$, and $LN$ denotes layer normalization~\cite{ba2016layer}, while $CAR$ denotes adaptive channel-wise recalibration to further enhance discriminative spatial-temporal-channel features, shown as:
\begin{equation}
F_{CAR}^{a/v} = F_{ST}^{a/v} \odot \sigma(W_2 \cdot ReLU(W_1 \cdot GAP(F_{ST}^{a/v}))).
\end{equation}
where $GAP$ represents the global average pooling operation applied across the temporal and spatial dimensions, $W_1 \in \mathbb{R}^{\frac{C}{r} \times C}$ and
$W_2 \in \mathbb{R}^{C \times \frac{C}{r}}$ are the learnable parameters of two fully connected layers, with $r$ denoting the reduction ratio (typically set to 16 in Ours), $\sigma$ and $ReLU$ are the $Sigmoid$ and $ReLU$ activation functions, and $\odot$ denotes the element-wise multiplication.

Then, the enhanced features are further correlated by Bidirectional Cross-modal Attention ($BCA$) to enforce consistent representations of $\bar{F}^{e}_{v}$ and $\bar{F}^{e}_{a}$ between visual and audio modalities in the different frequency domains, shown as:
\begin{equation}
\begin{gathered}
\bar{F}^{e}_{v} = f^{e}_{a\rightarrow v}(STC^{e}_{q}(\hat{F}_{FDED}^{v}), STC^{e}_{k}(\hat{F}_{FDED}^{a}) , \\ STC^{e}_{v}(\hat{F}_{FDED}^{a})),\\
\bar{F}^{e}_{a} = f^{e}_{v\rightarrow a}(STC^{e}_{q}(\hat{F}_{FDED}^{a}), STC^{e}_{k}(\hat{F}_{FDED}^{v}) , \\STC^{e}_{v}(\hat{F}_{FDED}^{v})),
\end{gathered}
\end{equation}
where $e$ denotes the process in the $e$-th expert, $f^{e}_{a\rightarrow v}$ and $f^{e}_{v\rightarrow a}$ denote the two directions in $BCA$ function.

The routing layer is responsible for dynamically integrating the corresponding experts based on the other modality. To be specific, after obtaining $\hat{F}_{FDED}^{v}$ and
$\hat{F}_{FDED}^{a}$, we first derive a compact global representation
through $STC$ operation
followed by global average pooling over the temporal
and spatial dimensions, and then, we employ an $MLP$ to produce the
weights by using the softmax function, shown as:
\begin{equation}
\begin{gathered}
W_{v} = SoftMax(MLP_a(
GAP(STC_a(\hat{F}_{FDED}^{a})))),\\
W_{a} = SoftMax(MLP_v(
GAP(STC_v(\hat{F}_{FDED}^{v})))),
\end{gathered}
\end{equation}
where $W_{a/v} \in \mathbb{R}^{N_e}$ and $W_{a/v}=[W^{e}_{a/v}]_{e=1}^{N_e}$. $W_v$ and $W_a$ are then applied to the expert outputs $\bar{F}_{v}^{e}$ and $\bar{F}_{a}^{e}$, respectively. This bidirectional incorporation with complementary modality not only enforces the modality consistency, but also restores suppressed features across modalities (e.g., visual cues guiding audio expert allocation and vice versa). 

Furthermore, to address the limitations of fixed top-$k$ routing~\cite{dai2022stablemoe}, which may lead to suboptimal expert utilization, we utilize the entropy of $W_{v}$ and $W_{a}$ to guide the dynamic selection of $k$ experts, shown as:
\begin{equation}
\begin{gathered}
E_{a/v} = -\sum_{e=1}^{N_e} W^{e}_{a/v} \log(W^{e}_{a/v} + \epsilon),\\
k_{a/v} = \left\lceil k_{\min} + (k_{\max} - k_{\min}) \times norm(E_{a/v})\right\rceil,
\end{gathered}
\end{equation}
where $k_{\min}$ and $k_{\max}$ are set to 1 and $N_e$, respectively,
$\epsilon$ is set to $10^{-8}$ for numerical stability, and
$E_{a/v}$ is normalized to the range $[0,1]$ through
$norm(E_{a/v})=E_{a/v}/\log N_e$.
Subsequently, we extract the top-$k_{a/v}$ scores and their indices
from $W_{a/v}$ for each sample, construct a dispatch mask to activate
the selected experts, and derive the normalized sparse weights
$\bar{W}_{a/v}$.
The selected probabilities are divided by their
sum, while the weights of unselected experts are set to zero.
Finally, the expert outputs are aggregated using the sparse
weights, shown as:
\begin{equation}
\begin{gathered}
\tilde F_{v},\tilde F_{a} = \sum_{e=1}^{N_e}\bar{W}^{e}_{v}\bar{F}^{e}_{v}, \sum_{e=1}^{N_e}\bar{W}^{e}_{a}\bar{F}^{e}_{a}.
\end{gathered}
\end{equation}

Ultimately, the aggregation yields fused features $\tilde F_{v}$ and $\tilde F_{a}$, which integrate semantic information and capture cross-modal correlations across the reconstructed frequency bands.
\begin{table}[t!]
\footnotesize
\caption{Comparison with state-of-the-art methods on the AVSS dataset. The best results are highlighted in \textbf{bold}, and the second-best are \underline{underlined}. $\mathcal{J}\&\mathcal{F}_m$ means the average value of $\mathcal{J}$ and $\mathcal{F}_m$. \# denotes two-stage methods; \textsuperscript{*} indicates results under the DDESeg evaluation protocol. }
\label{avsstable}
\centering
\resizebox{\columnwidth}{!}{%
\renewcommand{\arraystretch}{1.1}
\begin{tabular}{l c c ccc}
\toprule
\multirow{2}{*}{Method} &
\multirow{2}{*}{Reference} &
\multirow{2}{*}{Backbone} &
\multicolumn{3}{c}{AVSS} \\
\cmidrule(lr){4-6}
& & &
$\mathcal{J}$ &
$\mathcal{F}_m$ &
$\mathcal{J}\&\mathcal{F}_m$ \\
\midrule
\multicolumn{6}{l}{\textbf{\textit{Results on 224$\times$224 Image Size}}}\\
\midrule
TPAVI & ECCV'22 & PVT-v2 & 29.8 & 35.2 & 32.5 \\
CATR & ACMMM'23 & PVT-v2 & 32.8 & 38.5 & 35.7 \\
AVSegformer & AAAI'24 & PVT-v2 & 36.7 & 42.0 & 39.4 \\
BAVS & TMM'24 & Swin-Base & 33.6 & 37.5 & 35.6 \\
DiffusionAVS & TIP'25 & PVT-v2 & 38.1 & 43.0 & 40.6 \\
COMBO & CVPR'24 & PVT-v2 & 42.1 & 46.1 & 44.1 \\
AVSStone \# & ECCV'24 & Swin-Base & 48.5 & 53.2 & 50.9 \\
SelM & ACMMM'24 & PVT-v2 & 41.3 & 46.9 & 44.1 \\
AVSAC & TCSVT'25 & PVT-v2 & 37.0 & 42.4 & 39.7 \\
DDESeg\textsuperscript{*} & CVPR'25 & PVT-v2 &
\textbf{60.8} & \underline{70.6} & \underline{65.7} \\

ALOHA &
TOMM'25 &
CAFormer-B36 &
40.2 &
47.1 &
43.7 \\

VCT &
CVPR'25 &
Swin-Base &
47.9 &
52.9 &
50.4 \\

Crab &
CVPR'25 &
ViT-L/14 &
26.59 &
32.10 &
29.35 \\

\rowcolor{red!12}
\textbf{FAVS (Ours)} & -- & Swin-Base & 45.5 & 50.6 & 48.1 \\
\rowcolor{red!5}
\textbf{FAVS (Ours)}\textsuperscript{*} & -- & Swin-Base &
\underline{60.6} & \textbf{72.1} & \textbf{66.4} \\
\midrule
\multicolumn{6}{l}{\textbf{\textit{Results on 384$\times$384 Image Size}}}\\
\midrule
TeSo & ECCV'24 & Swin-Base & 39.0 & 45.1 & 42.1 \\
AVSBias & ACMMM'24 & Swin-Base & 44.4 & 49.9 & 47.2 \\
CQFormer & TIP'25 & PVT-v2 & 38.1 & 43.0 & 40.6 \\
\rowcolor{blue!5}
\textbf{FAVS (Ours)} & -- & Swin-Base &
\textbf{48.3} & \textbf{52.8} & \textbf{50.6} \\
\bottomrule
\end{tabular}%
}
\end{table}
\section{Experiments}
\subsection{Datasets}
We evaluate our model on three subsets derived from AVSBench~\cite{zhou2022audio, zhou2025audio}: Semi-supervised Single-source Sound segmentation (S4), Fully-supervised Multiple-sound Source Segmentation (MS3), and Fully-supervised AVSBench-Semantic (AVSS). S4 provides only initial-frame ground truth masks for binary segmentation of a single sounding object. In contrast, MS3 and AVSS target multi-source scenarios and offer frame-level supervision during training. While MS3 focuses on simultaneously segmenting multiple sounding objects, AVSS introduces the additional complexity of predicting semantic masks, requiring both localization and classification of sounding regions. Collectively, MS3 and AVSS present significantly more challenging and realistic audio-visual segmentation tasks than the S4.

\subsection{Training Details}
We conduct training and evaluation on S4 and MS3 using an NVIDIA GeForce RTX 4090 GPU, and on AVSS using an NVIDIA A6000 GPU, choosing the released Mask2Former~\cite{cheng2022masked} model with Swin-Base~\cite{liu2021swin} visual backbone pre-trained on ADE20k~\cite{zhou2017scene} as weight initialization, and all input frames are resized to 224 × 224 or 384 × 384. We also use VGGish~\cite{hershey2017cnn} pre-trained on Audioset~\cite{gemmeke2017audio} as the audio backbone to extract the audio feature from 96 × 64 log-Mel spectrograms. Besides, we adopt the standard transformer decoder, with $L = 3$ (i.e., a total of 9 layers) and $N_q = 100$ as the default. During training, the comprehensive training loss comprises classification loss and mask loss. The classification loss is formulated by a cross-entropy loss, denoted as $L_{cls} = L_{ce}$. The mask loss integrates the binary cross-entropy loss and the dice loss, and is depicted as $L_{mask} = L_{bce} + L_{dice}$. The final loss $L$ can be expressed as $L = \lambda_{cls}L_{cls} + \lambda_{mask}L_{mask}$, where $\lambda_{cls}$ and $\lambda_{mask}$ are set as 2 and 5. We trained the S4 and AVSS subsets for 100k and MS3 for 20k iterations, and all the models are trained using AdamW as the optimizer and a multi-step learning rate scheduler, with a batch size of 4, an initial learning rate of 1e-4 and weight decay of 0.05. On the RTX 4090, training with a batch size of four five-frame clips and an input resolution of $224\times224$ requires approximately 16 GiB of peak GPU memory. FAVS achieves an inference speed of 55.4 FPS.

\subsection{Evaluation Metrics}
Regarding evaluation metrics, we follow prior works~\cite{zhou2022audio, zhou2025audio, liu2025dynamic} to adopt $\mathcal{J}$ and $\mathcal{F}_m$ for performance measurement. $\mathcal{J}$ is the Jaccard index which computes the Intersection over Union (IoU) between predicted segmentation and ground truth, and $\mathcal{F}$ denotes the F-score which combines the precision and recall of the segmentation results. Our experiments report the mean Jaccard Index ($\mathcal{J}$) and mean F-score ($\mathcal{F}_m$) averaged across all test samples. For the prediction-threshold sensitivity analysis
on S4, 0.5 is used as the reference threshold. The corresponding
$\mathcal{J}$ and $\mathcal{F}_m$ are computed from masks binarized
at each fixed threshold, with $\beta^2=0.3$ for the F-score.
\begin{table}[!t]
\footnotesize
\centering
\setlength{\tabcolsep}{2.5pt}
\renewcommand{\arraystretch}{1.1}
\caption{Ablation Study of the Two Proposed Modules Included
in FAVS. Params includes frozen parameters;
FLOPs are per five-frame clip. Cost increments in the text
are calculated before rounding.}
\label{tab:results}
\begin{tabular}{lcccccc}
\toprule
\multirow{2}{*}{Method} &
\multicolumn{2}{c}{S4} &
\multicolumn{2}{c}{MS3} &
\multirow{2}{*}{\shortstack{
Params\\
(M)}} &
\multirow{2}{*}{\shortstack{
FLOPs\\
(G)}} \\
\cmidrule(lr){2-3}\cmidrule(lr){4-5}
& $\mathcal{J}$ & $\mathcal{F}_m$
& $\mathcal{J}$ & $\mathcal{F}_m$ & & \\
\midrule
Baseline &
84.3 & 90.9 & 67.6 & 75.3 &
188.1 &
184.5 \\

\rowcolor{red!5}
(+) FDED &
{+0.5} & +0.5 &
{+1.4} & +1.5 &
192.2 &
203.8 \\

\rowcolor{red!12}
\textbf{(+) FDED (+) SCMC} &
\textbf{+1.3} & \textbf{+1.3} &
\textbf{+3.5} & \textbf{+3.4} &
\textbf{199.5} &
\textbf{210.0} \\
\bottomrule
\end{tabular}
\end{table}
\subsection{Comparisons with State-of-the-art Methods}
We compare our method with previous state-of-the-art approaches TPAVI~\cite{zhou2022audio, zhou2025audio}, AQFormer~\cite{huang2023discovering}, CATR~\cite{li2023catr}, AVSC~\cite{liu2023audio1}, ECMVAE~\cite{mao2023multimodal}, BAVS~\cite{liu2024bavs}, AVSBG~\cite{hao2024improving}, GAVS~\cite{wang2024prompting}, AVSegformer~\cite{gao2024avsegformer}, UFE~\cite{liu2024audio}, COMBO~\cite{yang2024cooperation}, AVSStone~\cite{ma2024stepping}, SelM~\cite{li2024selm}, TeSo~\cite{wang2024can}, AVSBias~\cite{sun2024unveiling}, DiffusionAVS~\cite{mao2025contrastive}, AVSAC~\cite{chen2025bootstrapping}, CQFormer~\cite{10979212} , DDESeg~\cite{liu2025dynamic}, ALOHA~\cite{zhou2025aloha}, METTLE~\cite{zhou2025mettle}, Crab~\cite{du2025crab} and VCT~\cite{huang2025revisiting}   on S4, MS3 and AVSS datasets using transformer-based visual backbones. 

\subsubsection{Quantitative Evaluation.}
Table~\ref{AVSexp} shows the quantitative comparisons between our
model and other recent methods on the S4 and MS3 datasets.
We can see that our model achieves strong
performance against these methods. Specifically, our model
surpasses COMBO by 0.9 $\mathcal{J}$ and
0.3 $\mathcal{F}_m$ for S4 and AVSStone by
3.8 $\mathcal{J}$ and 1.1 $\mathcal{F}_m$ for MS3, highlighting
its robustness across both single-source and multi-source tasks.
With the same Swin-Base backbone and
$224\times224$ inputs, FAVS also surpasses VCT on MS3 by
3.6 percentage points in $\mathcal{J}$ and 1.5 points in
$\mathcal{J}\&\mathcal{F}_m$.
Additionally, Table~\ref{avsstable} offers a detailed comparison
of our method with others on the AVSBench-semantic datasets.
Notably, Ours also achieves significant results on the AVSS,
achieving 3.9 $\mathcal{J}$ improvements and significant
$\mathcal{F}_m$ enhancements of 2.9 over
AVSBias at an input resolution of $384\times384$.
FAVS further achieves higher $\mathcal{J}$ than
ALOHA, METTLE, and the large multimodal model
Crab~\cite{du2025crab} on their reported subsets.
The experimental results show that our model
achieves substantial improvements on the more
challenging datasets (MS3 and AVSS), demonstrating its effectiveness
in complex multi-source scenes.

\subsubsection{Qualitative Evaluation.} 
\begin{figure*}[ht]
  \centering
  \includegraphics[width=\linewidth, height=0.5\linewidth]{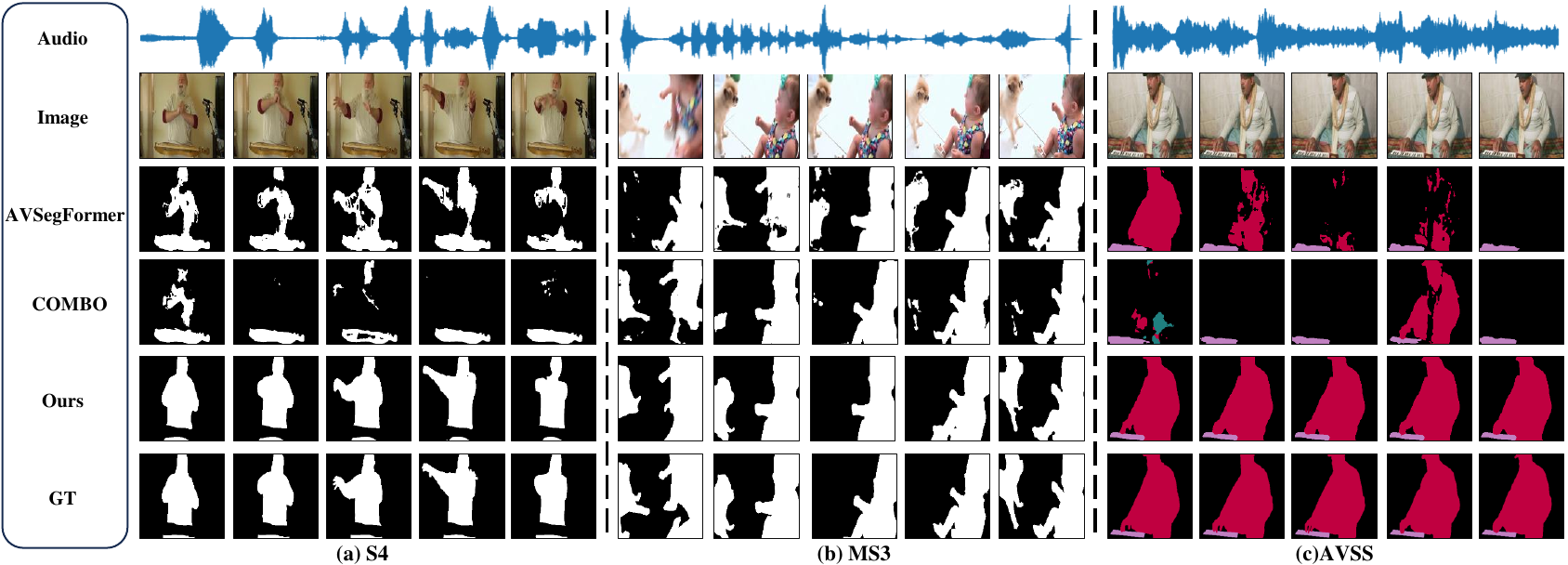}
  \caption{Qualitative comparison between AVSegFormer, COMBO and Ours on the three subsets of AVSBench dataset. (a) demonstrates the examples derived from S4, (b) demonstrates the examples derived from MS3, and (c) demonstrates the examples derived from AVSS.}
  \vspace{-10pt}
  \label{avs}
\end{figure*}

To showcase the visual effect of our model's segmentation effects, we present the visualization results of our model compared with those of AVSegformer and COMBO in Fig.~\ref{avs}. As shown in Fig.~\ref{avs} (a), the man and instrument can not be segmented well by previous methods, especially in the details of the edges. Meanwhile, in the multi-source scenes (as seen in Fig.~\ref{avs} (b) and (c)), our model accurately segments multiple simultaneously sounding objects, while competing approaches suffer from incomplete segmentation. These results demonstrate that our model delivers precise segmentation for enhanced boundary delineation, while also exhibiting strong semantic understanding that enables effective recognition of the correct sound source in complex scenes.

\subsection{Ablation Experiments}
In order to verify the effectiveness of each key design, we perform ablation studies in S4 and MS3 datasets. Notably, all experiments used the Swin-Base backbone as the visual encoder and all input frames are resized to 224 × 224.

To evaluate the contribution of FDED and SCMC modules to the
overall model performance, we add each module to the baseline.
As presented in Table~\ref{tab:results},
FDED improves segmentation through modality-specific
frequency refinement, while SCMC provides further gains through
cross-modal interaction. The larger gains on MS3 suggest that
selectively refining frequency information and establishing
reliable audio-visual associations are particularly useful
when multiple sound sources coexist.

The computational breakdown shows that SCMC
introduces more additional parameters but fewer additional FLOPs
than FDED. Its multi-expert interaction therefore improves
segmentation with a relatively small increase in computation
over the FDED-only configuration.
\begin{figure}[!t]
  \centering
  \includegraphics[width=\linewidth, height=0.63\linewidth]{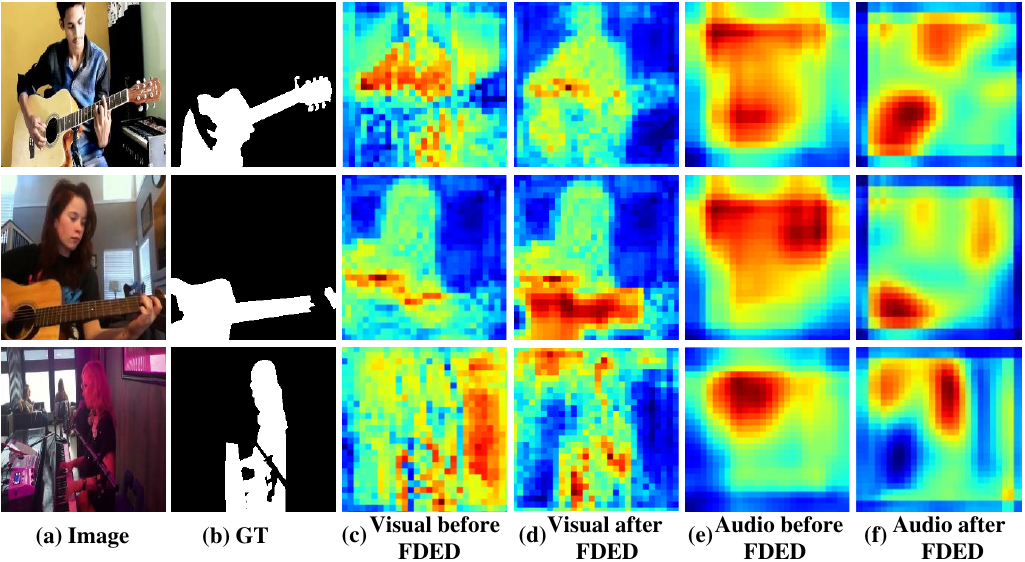}
  \caption{Feature maps visualizations in FDED presented in sequence from left to right. (a) raw images, (b) ground truth masks, (c) visual feature maps before processing by FDED, (d) visual feature maps after processing by FDED, (e) audio feature maps before processing by FDED, and (f) audio feature maps after processing by FDED.}
  \label{FDED}
  \vspace{-15pt}
\end{figure}

\subsubsection{Visualization on FDED.} To prove the effects of FDED module from a more intuitive perspective, we illustrate the visual and audio feature maps on FDED module. As shown in Fig.~\ref{FDED}, we can see a progressive enhancement on edge information within the visual feature maps compared Fig.~\ref{FDED}(c) with (d), reflecting FDED's effective reinforcement of high-frequency components critical for capturing structural details. Meanwhile, Fig.~\ref{FDED}(e) and (f) indicate a transition where audio features become concentrated on sounding objects, suggesting that FDED module suppresses high-frequency noise in the audio while emphasizing low-to-middle frequency bands that encode semantic audio information.
\begin{table}[!t]
\centering
\renewcommand{\arraystretch}{1.1}
\caption{Impact of frequency band number in the FDED module.}
\begin{tabular}{ccccc}
\toprule
\multirow{2}{*}{\textbf{Frequency bands}} & 
\multicolumn{2}{c}{\textbf{S4}} & 
\multicolumn{2}{c}{\textbf{MS3}} \\
\cmidrule(lr){2-3} \cmidrule(lr){4-5}
& $\mathcal{J}$ & $\mathcal{F}_m$ & $\mathcal{J}$ & $\mathcal{F}_m$ \\
\midrule
Two & 84.5 & 91.0 & 68.6 & 76.2 \\
\textbf{Three} & \textbf{84.8} & \textbf{91.4} & \textbf{69.0} & \textbf{76.8} \\
Four & 84.3 & 91.1 & 68.4 & 75.8 \\
\bottomrule
\end{tabular}
\label{tab:frequency_bands}
\end{table}
\begin{figure}[!t]
  \centering
  \includegraphics[width=\linewidth, height=0.63\linewidth]{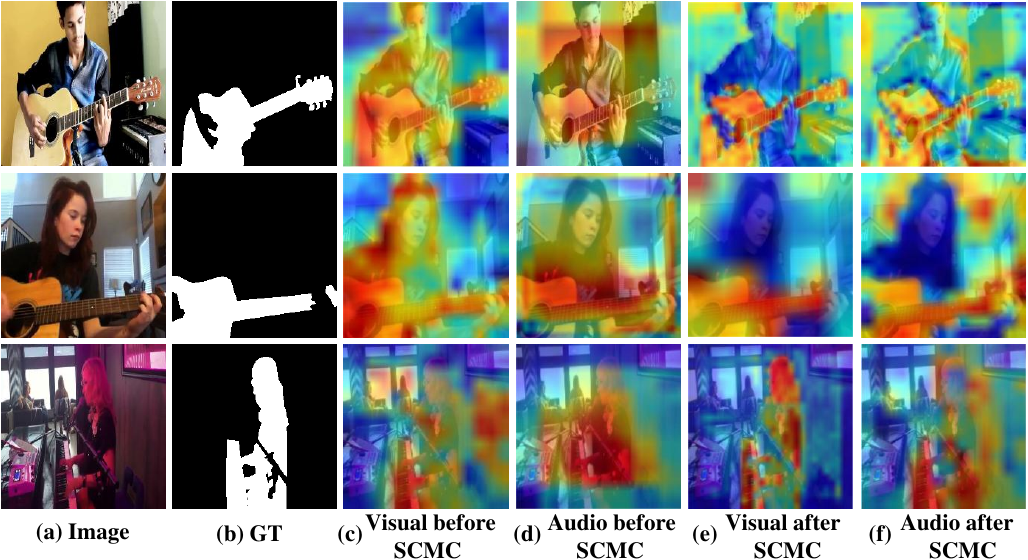}
  \caption{Feature maps visualizations in SCMC presented in sequence from left to right. (a) raw images, (b) ground truth masks, (c) visual feature maps before processing by SCMC, (d) audio feature maps before processing by SCMC, (e) visual feature maps after processing by SCMC, and (f) audio feature maps after processing by SCMC.}
  \label{SCMC1}
\end{figure}
\subsubsection{Impact of Frequency Band Number in FDED.}
To comprehensively evaluate the impact of frequency band number on FDED module performance, we conduct residual-based iterative decomposition by testing different numbers of frequency bands along with their corresponding frequency thresholds $\tau$. At each iteration, we determine a frequency threshold $\tau$ based on the current remaining frequency spectrum, uniformly partitioning it into distinct bands. Specifically, without considering the residual band, we compare three decomposition granularities with corresponding $\tau$ values: (1) two bands (high and low frequencies, $\tau = 0.5$), (2) three bands (high, middle, and low frequencies, $\tau = 0.7, 0.3$, \textbf{Ours}), and (3) four bands (high, sub-high, middle, and low frequencies, $\tau = 0.8, 0.5, 0.3$). Note that this ablation focuses solely on the FDED module, excluding the SCMC module.

As illustrated in Table~\ref{tab:frequency_bands}, the performance improves from two to three frequency bands, achieving peak values of 84.8 $\mathcal{J}$ and 91.4 $\mathcal{F}_m$ on S4, and 69.0 $\mathcal{J}$ and 76.8 $\mathcal{F}_m$ on MS3. Further increasing to four bands leads to a slight performance drop. These results numerically confirm that an appropriate division of the spectrum into high, middle, and low-frequency bands facilitates effective separation and targeted modality-specific enhancement—especially of high-frequency components—leading to significantly improved segmentation performance.

When the frequency decomposition is insufficient (e.g., only two bands), modality-specific features remain intermixed within the same frequency bands, causing modality-specific enhancements to be less effective and thus limiting overall performance improvement. Additionally, excessively fine-grained decomposition (e.g., four bands) can introduce redundant information or dilute the model’s focus on critical frequency components, thereby weakening feature enhancement effectiveness, especially in complex, multisource scenarios, resulting in a performance drop.

\begin{figure}[!t]
  \centering
  \includegraphics[width=\linewidth, height=0.63\linewidth]{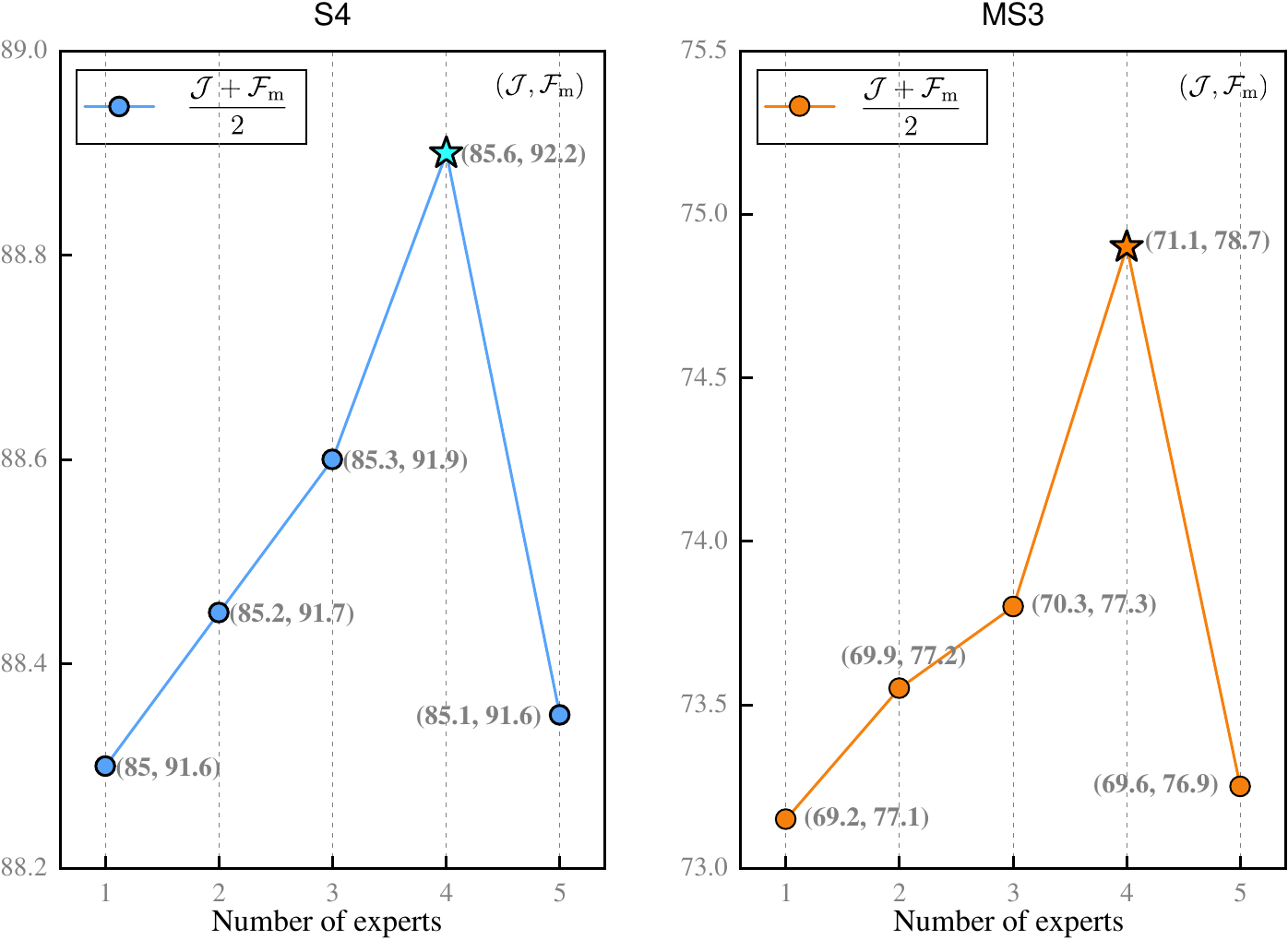}
  \caption{Effect of the number of experts in SCMC on S4 and MS3 datasets. The y-axis reports the average score $(\mathcal{J}+\mathcal{F}_m)/2$. Performance increases steadily from 1 to 4 experts and drops when using 5 experts, indicating that 4 experts provide the best balance between specialization and redundancy.}
  \label{MOE}
\end{figure}
\begin{figure}[!t]
  \centering
  \includegraphics[width=\linewidth, height=0.6\linewidth]{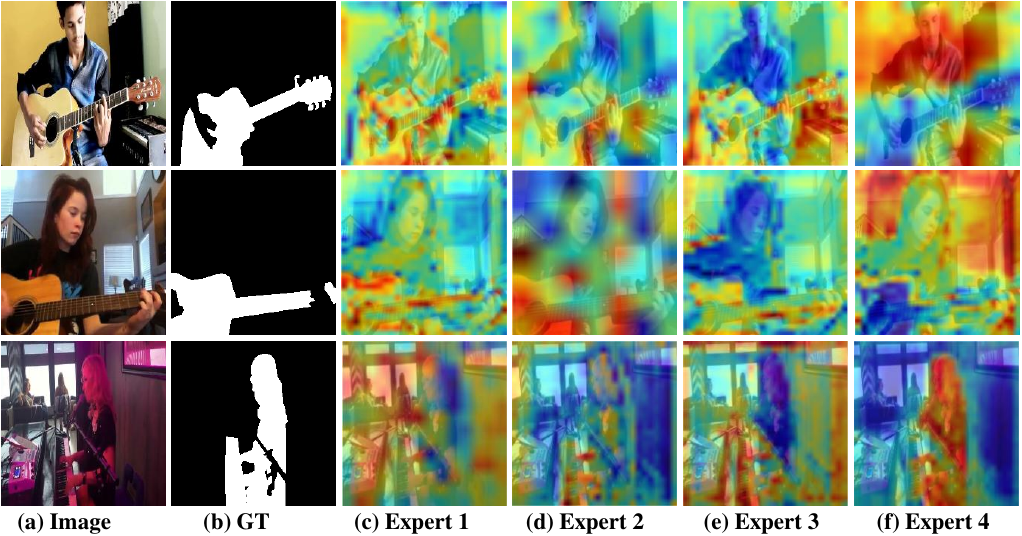}
  \caption{The visual feature maps visualizations of the outputs from the four attention experts in SCMC. (a) raw images, (b) ground truth masks, (c)-(f) the visual feature maps from expert 1 to expert 4.}
  \label{SCMC2}
  \vspace{-5pt}
\end{figure}

\subsubsection{Contributions of Frequency Bands
and High-Frequency Processing.}
To examine the contributions of individual
frequency bands and their processing, we conduct the ablations
in Table~\ref{tab:band_contribution}. Removing high-frequency
audio causes a small performance decline, whereas retaining
it without selective processing is more detrimental on MS3.
This supports preserving useful high-frequency cues while
selectively suppressing irrelevant responses before recomposition.

\begin{table}[!t]
\footnotesize

\centering
\setlength{\tabcolsep}{2pt}
\renewcommand{\arraystretch}{1.1}
\caption{Ablation of frequency bands and processing in FDED.
A/V: audio/visual; H/M/L: high/mid/low bands;
P: high-frequency processing. Bold: full-model scores (\%);
other entries: changes from the full model in percentage points.}
\label{tab:band_contribution}
\begin{tabular}{ccccc ccc ccc}
\toprule
\multirow{2}{*}{ID} &
\multirow{2}{*}{H} &
\multirow{2}{*}{M} &
\multirow{2}{*}{L} &
\multirow{2}{*}{P} &
\multicolumn{3}{c}{S4} &
\multicolumn{3}{c}{MS3} \\
\cmidrule(lr){6-8}\cmidrule(lr){9-11}
& & & & &
$\mathcal{J}$ &
$\mathcal{F}_m$ &
$\mathcal{J}\&\mathcal{F}_m$ &
$\mathcal{J}$ &
$\mathcal{F}_m$ &
$\mathcal{J}\&\mathcal{F}_m$ \\
\midrule

\rowcolor{red!12}
A1 & $\checkmark$ & $\checkmark$ & $\checkmark$ & $\checkmark$ &
\textbf{85.65} & \textbf{92.16} & \textbf{88.91} &
\textbf{71.12} & \textbf{78.72} & \textbf{74.92} \\

A2 & $\checkmark$ & $\checkmark$ & $\checkmark$ & $\times$ &
0.00 & 0.00 & 0.00 &
$-1.77$ & $-1.62$ & $-1.70$ \\

A3 & $\times$ & $\checkmark$ & $\checkmark$ & $\times$ &
$-0.09$ & $-0.05$ & $-0.07$ &
$-0.14$ & $-0.28$ & $-0.21$ \\

A4 & $\checkmark$ & $\times$ & $\checkmark$ & $\checkmark$ &
$-0.04$ & $-0.02$ & $-0.03$ &
$-0.10$ & $-0.47$ & $-0.29$ \\

A5 & $\checkmark$ & $\checkmark$ & $\times$ & $\checkmark$ &
$-0.04$ & $-0.02$ & $-0.03$ &
$-0.86$ & $-0.68$ & $-0.77$ \\

\midrule

\rowcolor{red!12}
V1 & $\checkmark$ & $\checkmark$ & $\checkmark$ & $\checkmark$ &
\textbf{85.65} & \textbf{92.16} & \textbf{88.91} &
\textbf{71.12} & \textbf{78.72} & \textbf{74.92} \\

V2 & $\checkmark$ & $\checkmark$ & $\checkmark$ & $\times$ &
$-0.65$ & $-0.36$ & $-0.51$ &
$-0.56$ & $-0.69$ & $-0.63$ \\

V3 & $\times$ & $\checkmark$ & $\checkmark$ & $\times$ &
$-0.82$ & $-0.39$ & $-0.61$ &
$-0.56$ & $-0.40$ & $-0.48$ \\

V4 & $\checkmark$ & $\times$ & $\checkmark$ & $\checkmark$ &
$-0.22$ & $-0.17$ & $-0.20$ &
$-0.72$ & $-0.77$ & $-0.75$ \\

V5 & $\checkmark$ & $\checkmark$ & $\times$ & $\checkmark$ &
$-0.08$ & $-0.02$ & $-0.05$ &
$-0.72$ & $-0.72$ & $-0.72$ \\

\bottomrule
\end{tabular}
\end{table}
Removing visual high frequencies or their
processing also degrades performance, supporting their role
in preserving structural information for segmentation.
Middle- and low-frequency bands provide complementary
information, with low-frequency audio having a greater
influence on MS3. Together, these results support retaining
information across bands and applying modality-specific
refinement, particularly in multi-source scenes.

\begin{table}[!t]
\footnotesize

\centering
\setlength{\tabcolsep}{2.5pt}
\renewcommand{\arraystretch}{1.1}
\caption{Sensitivity to frequency boundaries in FDED.
Bold: default scores (\%); other entries: changes from
the default in percentage points.}
\label{tab:frequency_thresholds}
\begin{tabular}{cc ccc ccc}
\toprule
\multirow{2}{*}{$\tau_1$} &
\multirow{2}{*}{$\tau_2$} &
\multicolumn{3}{c}{S4} &
\multicolumn{3}{c}{MS3} \\
\cmidrule(lr){3-5}\cmidrule(lr){6-8}
& &
$\mathcal{J}$ &
$\mathcal{F}_m$ &
$\mathcal{J}\&\mathcal{F}_m$ &
$\mathcal{J}$ &
$\mathcal{F}_m$ &
$\mathcal{J}\&\mathcal{F}_m$ \\
\midrule

\rowcolor{red!12}
\textbf{0.7} & \textbf{0.3} &
\textbf{85.65} & \textbf{92.16} & \textbf{88.91} &
\textbf{71.12} & \textbf{78.72} & \textbf{74.92} \\

0.6 & 0.3 &
$-0.12$ & $-0.06$ & $-0.09$ &
$-0.10$ & $-0.13$ & $-0.12$ \\

0.8 & 0.3 &
$-0.07$ & $-0.03$ & $-0.05$ &
$+0.06$ & $-0.03$ & $+0.02$ \\

0.7 & 0.2 &
$-0.04$ & $+0.02$ & $-0.01$ &
$-0.16$ & $-0.09$ & $-0.13$ \\

0.7 & 0.4 &
$-0.16$ & $-0.10$ & $-0.13$ &
$-0.28$ & $-0.25$ & $-0.27$ \\

\bottomrule
\end{tabular}
\end{table}
\subsubsection{Sensitivity to Frequency Boundaries.}
Building on the frequency-band number study,
we further examine the sensitivity of the three-band configuration
to its frequency boundaries. The high/middle boundary $\tau_1$
and middle/low boundary $\tau_2$ are varied separately around
their default values, as shown in
Table~\ref{tab:frequency_thresholds}. Performance varies only
slightly within the tested range, although increasing $\tau_2$
produces a more noticeable change on MS3. Together with the
module ablation, these results suggest that the gains mainly
stem from the frequency-domain processing in FDED, with limited
dependence on the precise boundary values within this range.

\subsubsection{Visualization on SCMC.} To provide a more perceptual perspective for evaluating how SCMC reinforces consistent representations across the modalities, we employ Grad-CAM~\cite{selvaraju2017grad} to present visualizations of the visual and audio feature maps generated by SCMC. As shown in Fig.~\ref{SCMC1}, before applying the SCMC module, the visual and audio feature maps (Fig.~\ref{SCMC1}(c) and (d)) exhibit pronounced spatial inconsistencies, stemming from their unique modality-oriented emphasis and insufficient correlation. In contrast, after SCMC module, the visual and audio feature maps (Fig.~\ref{SCMC1}(e) and (f)) show improved consistency, highlighting the module's capability to promote coherent and discriminative cross-modal synergy.

\subsubsection{Impact of Different Expert Numbers.} As shown in Fig.~\ref{MOE}, we first investigate the impact of the number of experts in SCMC module. It can be seen that increasing the number of experts initially improves model performance, peaking at 4 experts before declining with further additions. These results numerically confirm that the design of experts is effective, while excessive experts may cause redundancy and routing mechanism risks, as they may overlap with similar cues, impairing integration efficiency and synergy.

\begin{figure}[!t]
  \centering
  \includegraphics[width=0.85\linewidth]{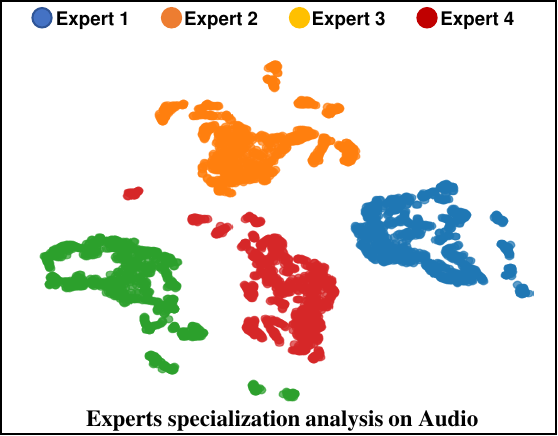}
  \caption{The visualization presents experts' ability to capture unique representations of audio features. }
  \label{SCMC3}
\end{figure}
\begin{figure}[!t]
  \centering
  \includegraphics[width=0.9\linewidth, height=0.7\linewidth]{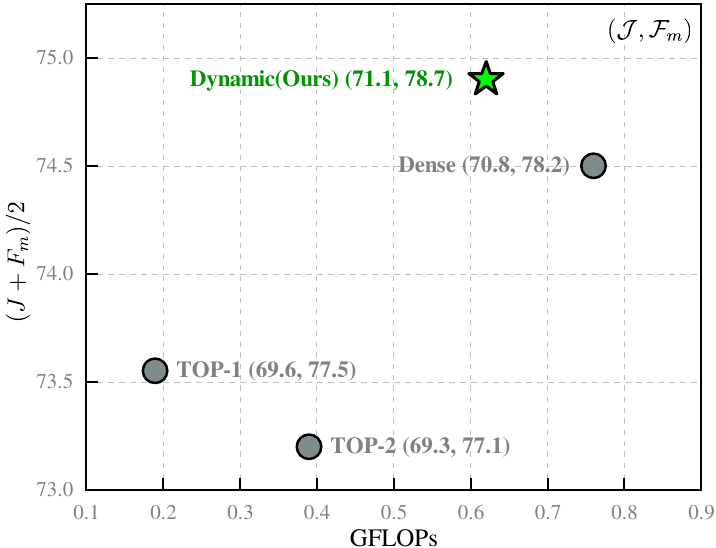}
  \caption{The visualization presents the performance and computational efficiency of routing mechanisms. The x-axis denotes GFLOPs and the y-axis denotes the average score $(\mathcal{J}+\mathcal{F}_m)/2$. Dynamic routing achieves a more favorable trade-off than fixed sparse routing while avoiding the highest computational cost of dense routing.}
  \vspace{-10pt}
  \label{GFLOPS}
\end{figure}
\subsubsection{Experts specialization analysis.} To deliver a more intuitive angle for analyzing how each attention expert learns to specialize in capturing corresponding features in distinct frequency bands, we present visualizations of the visual feature maps generated by each expert. As shown in Fig.~\ref{SCMC2}, SCMC module enables each expert to specialize in various aspects and capture complementary correlations. Then, these experts leverage weights derived from complementary modalities to selectively amplify salient features and suppress misleading ones, promoting consistent cross-modal representations. Meanwhile, to further elucidate the specialization of each attention expert, we employ t-SNE~\cite{maaten2008visualizing} dimensionality reduction to visualize the audio features output by individual experts, as shown in Fig.~\ref{SCMC3}. The clear clustering underscores the experts' ability to capture unique representations, highlighting their complementary contributions to cross-modal consistency.

\subsubsection{Impact of Dynamic Routing mechanism in SCMC.} To assess the effectiveness of the dynamic routing mechanism in SCMC, we conduct an experiment by comparing it with top-1, top-2, and dense routing mechanisms. As shown in Fig.~\ref{GFLOPS}, the results reveal the dynamic routing's superiority: on MS3, it achieves $\mathcal{J}$ of 71.1 and $\mathcal{F}_m$ of 78.7, outperforming others by 0.3–1.8 $\mathcal{J}$ and 0.5–1.6 $\mathcal{F}_m$. The dynamic router activates an average of
2.8 experts per input during inference. Meanwhile, the $\textbf{FLOPs}$ denotes the average FLOPs in SCMC during testing, and the dynamic routing mechanism delivers superior segmentation performance while incurring lower FLOPs than the dense routing approach. These findings affirm the dynamic router's adaptive efficiency in promoting consistency across modalities, as well as enhancing segmentation performance over fixed-sparse or dense methods.

\begin{table}[!t]
\footnotesize

\centering
\caption{Prediction-threshold sensitivity on S4.
Parentheses indicate changes from 0.5 in percentage points.}
\label{tab:prediction_threshold}
\setlength{\tabcolsep}{8pt}
\renewcommand{\arraystretch}{1.1}
\begin{tabular}{ccc}
\toprule
Threshold & $\mathcal{J}$ (\%) & $\mathcal{F}_m$ (\%) \\
\midrule
0.4 & 85.58 ($-0.02$) & 91.37 ($-0.83$) \\
\rowcolor{red!12}
\textbf{0.5} & \textbf{85.60} & \textbf{92.20} \\
0.6 & 84.93 ($-0.67$) & 92.60 ($+0.40$) \\
\bottomrule
\end{tabular}
\end{table}
\subsubsection{Sensitivity to the Prediction Threshold.}
To examine the influence of the prediction
threshold, we consider 0.4, 0.5, and 0.6 on S4, using 0.5 as
the reference setting. Table~\ref{tab:prediction_threshold} presents the corresponding
$\mathcal{J}$ and $\mathcal{F}_m$. Lowering the threshold
to 0.4 leaves $\mathcal{J}$ nearly unchanged but reduces
$\mathcal{F}_m$, consistent with including additional
low-confidence foreground pixels. Increasing it to 0.6
produces more conservative foreground masks, with lower
$\mathcal{J}$ but higher $\mathcal{F}_m$. These trends are
consistent with a trade-off between foreground coverage
and precision when adjusting the prediction threshold.
\begin{table}[!t]
\footnotesize

\centering
\caption{Comparison of decomposition strategies.
Parentheses indicate changes from the proposed method
in percentage points.}
\label{tab:decomposition_comparison}
\setlength{\tabcolsep}{3pt}
\renewcommand{\arraystretch}{1.1}
\resizebox{\columnwidth}{!}{%
\begin{tabular}{lcccc}
\toprule
\multirow{2}{*}{Decomposition}
& \multicolumn{2}{c}{S4}
& \multicolumn{2}{c}{MS3} \\
\cmidrule(lr){2-3}\cmidrule(lr){4-5}
& $\mathcal{J}$ & $\mathcal{F}_m$
& $\mathcal{J}$ & $\mathcal{F}_m$ \\
\midrule
Haar DWT
& 85.24 ($-0.36$) & 91.96 ($-0.24$)
& 70.62 ($-0.48$) & 78.35 ($-0.35$) \\
\rowcolor{red!12}
\textbf{Fourier (Ours)}
& \textbf{85.60} & \textbf{92.20}
& \textbf{71.10} & \textbf{78.70} \\
\bottomrule
\end{tabular}%
}

\end{table}
\subsubsection{Comparison of Decomposition Strategies.}
To examine the influence of the decomposition
strategy, we compare the proposed Fourier-based decomposition
with a Haar discrete wavelet transform (DWT), retaining the
same backbone, modality-specific processing, SCMC module,
and training and evaluation settings.
As shown in Table~\ref{tab:decomposition_comparison},
the two variants achieve comparable performance, suggesting
that FAVS can also accommodate wavelet-based decomposition.
Both approaches separate feature components for subsequent
refinement and recomposition. We retain the Fourier-based
formulation because its explicit frequency boundaries provide
a direct way to organize the high-, middle-, and low-frequency
components used in FDED.
\subsubsection{Robustness to Audio Interference.}
To evaluate the contribution of audio refinement
in FDED under acoustic interference, we compare the full model
with a variant without this operation, using identical perturbed
inputs without additional training. Brownian and moving-train
noise are added to the entire audio clip at SNRs of 20, 10,
and 0 dB. We also mix unrelated audio with the original
target sound at an SIR of 10 dB.
\begin{table}[!t]
\footnotesize

\centering
\caption{Audio robustness in $\mathcal{J}$ (\%).
$P$: audio refinement in FDED.
Unrelated audio uses SIR instead of SNR.}
\label{tab:audio_robustness}
\setlength{\tabcolsep}{3pt}
\renewcommand{\arraystretch}{1.1}
\begin{tabular}{lccccc}
\toprule
\multirow{2}{*}{Noise}
& \multirow{2}{*}{\shortstack{SNR\\(dB)}}
& \multicolumn{2}{c}{S4}
& \multicolumn{2}{c}{MS3} \\
\cmidrule(lr){3-4}\cmidrule(lr){5-6}
& & Full & w/o $P$ & Full & w/o $P$ \\
\midrule
\rowcolor{red!12}
Clean & -- & 85.61 & 85.60 & 71.12 & 69.35 \\
\midrule
\multirow{3}{*}{Brownian}
& 20 & 85.61 & 85.61 & 71.13 & 69.36 \\
& 10 & 85.61 & 85.61 & 71.33 & 69.50 \\
& 0  & 85.65 & 85.63 & 71.16 & 69.59 \\
\midrule
\multirow{3}{*}{Moving train}
& 20 & 85.63 & 85.62 & 71.00 & 69.69 \\
& 10 & 85.64 & 85.64 & 70.06 & 68.81 \\
& 0  & 85.65 & 85.65 & 68.68 & 67.72 \\
\midrule
Unrelated audio
& 10 & 85.62 & 85.62 & 68.69 & 67.71 \\
\bottomrule
\end{tabular}
\end{table}
As shown in Table~\ref{tab:audio_robustness},
performance on S4 remains nearly unchanged.
On MS3, decreasing SNR causes greater degradation under
moving-train noise, whereas performance remains comparatively
stable under Brownian noise. The full model retains higher
$\mathcal{J}$ than the ablated variant across these conditions
and under unrelated audio interference. These retained
advantages support the contribution of audio refinement
to maintaining target segmentation under interference,
consistent with its intended role in suppressing irrelevant
audio responses before cross-modal interaction.
\section{Limitation and Future Work}
While FAVS demonstrates robust performance in challenging scenarios within the AVSBench datasets, it is predominantly trained and evaluated on datasets featuring predefined object categories, thereby constraining its generalization to novel object-sound pairings beyond the training distribution. This inherent limitation arises from its dependence on fully supervised learning paradigms applied to fixed datasets with a limited number of object-sound categories, predisposing the model to overfitting on familiar categories and impeding its handling of semantic and distributional discrepancies in unfamiliar contexts, ultimately hindering efficacy in real-world applications. To address this, future enhancements could incorporate advancements in large-scale multimodal foundation models, such as large multimodal models (LMMs) or open-vocabulary architectures, which exploit extensive pretraining corpora to enable emergent generalization through techniques like prompt engineering or adapter-based fine-tuning, thus bolstering adaptability across varied audio-visual environments. Our frequency-boundary analysis covers a local
range around the default setting, and the robustness tests
consider a limited set of noise types and interference levels.
Performance under broader frequency distributions and more
diverse acoustic conditions remains to be investigated.
Future work could explore input-dependent frequency selection
to better accommodate variations in audio-visual features.

\section{Conclusion}
We introduce a novel Frequency-Aware Audio-Visual Segmentation framework, termed FAVS, which achieves state-of-the-art performance on three benchmark datasets (S4, MS3, and AVSS). Contrary to previous methodologies that only consider temporal or spatial-modal relationships individually, our approach explores a deeper and previously overlooked issue—the inherent frequency-domain contradictions between audio and visual modalities. To address this challenge, we propose two innovative modules: the Frequency-Domain Enhanced Decomposer (FDED) module and the Synergistic Cross-Modal Consistency (SCMC) module. Extensive experimental results and significant qualitative visualizations demonstrate the effectiveness and superiority of our proposed framework. We hope our study can inspire further research into frequency-domain analysis for audio-visual segmentation.


\bibliographystyle{IEEEtran}
\bibliography{TIP}

@inproceedings{zhou2022audio,
  title={Audio--visual segmentation},
  author={Zhou, Jinxing and Wang, Jianyuan and Zhang, Jiayi and Sun, Weixuan and Zhang, Jing and Birchfield, Stan and Guo, Dan and Kong, Lingpeng and Wang, Meng and Zhong, Yiran},
  booktitle={European Conference on Computer Vision},
  pages={386--403},
  year={2022},
  organization={Springer}
}

@inproceedings{gao2024avsegformer,
  title={Avsegformer: Audio-visual segmentation with transformer},
  author={Gao, Shengyi and Chen, Zhe and Chen, Guo and Wang, Wenhai and Lu, Tong},
  booktitle={Proceedings of the AAAI conference on artificial intelligence},
  volume={38},
  pages={12155--12163},
  year={2024}
}

@inproceedings{ma2024stepping,
  title={Stepping stones: a progressive training strategy for audio-visual semantic segmentation},
  author={Ma, Juncheng and Sun, Peiwen and Wang, Yaoting and Hu, Di},
  booktitle={European Conference on Computer Vision},
  pages={311--327},
  year={2024},
  organization={Springer}
}

@inproceedings{wang2024ref,
  title={Ref-avs: Refer and segment objects in audio-visual scenes},
  author={Wang, Yaoting and Sun, Peiwen and Zhou, Dongzhan and Li, Guangyao and Zhang, Honggang and Hu, Di},
  booktitle={European Conference on Computer Vision},
  pages={196--213},
  year={2024},
  organization={Springer}
}

@article{huang2023discovering,
  title={Discovering sounding objects by audio queries for audio visual segmentation},
  author={Huang, Shaofei and Li, Han and Wang, Yuqing and Zhu, Hongji and Dai, Jiao and Han, Jizhong and Rong, Wenge and Liu, Si},
  journal={arXiv preprint arXiv:2309.09501},
  year={2023}
}

@inproceedings{li2023catr,
  title={Catr: Combinatorial-dependence audio-queried transformer for audio-visual video segmentation},
  author={Li, Kexin and Yang, Zongxin and Chen, Lei and Yang, Yi and Xiao, Jun},
  booktitle={Proceedings of the 31st ACM international conference on multimedia},
  pages={1485--1494},
  year={2023}
}

@article{liu2023audio,
  title={Audio-aware query-enhanced transformer for audio-visual segmentation},
  author={Liu, Jinxiang and Ju, Chen and Ma, Chaofan and Wang, Yanfeng and Wang, Yu and Zhang, Ya},
  journal={arXiv preprint arXiv:2307.13236},
  year={2023}
}

@inproceedings{ling2024transavs,
  title={TransAVS: End-to-End Audio-Visual Segmentation with Transformer},
  author={Ling, Yuhang and Li, Yuxi and Gan, Zhenye and Zhang, Jiangning and Chi, Mingmin and Wang, Yabiao},
  booktitle={ICASSP 2024-2024 IEEE International Conference on Acoustics, Speech and Signal Processing (ICASSP)},
  pages={7845--7849},
  year={2024},
  organization={IEEE}
}

@inproceedings{hao2024improving,
  title={Improving audio-visual segmentation with bidirectional generation},
  author={Hao, Dawei and Mao, Yuxin and He, Bowen and Han, Xiaodong and Dai, Yuchao and Zhong, Yiran},
  booktitle={Proceedings of the AAAI conference on artificial intelligence},
  volume={38},
  pages={2067--2075},
  year={2024}
}

@inproceedings{chollet2017xception,
  title={Xception: Deep learning with depthwise separable convolutions},
  author={Chollet, Fran{\c{c}}ois},
  booktitle={Proceedings of the IEEE conference on computer vision and pattern recognition},
  pages={1251--1258},
  year={2017}
}

@inproceedings{liu2021swin,
  title={Swin transformer: Hierarchical vision transformer using shifted windows},
  author={Liu, Ze and Lin, Yutong and Cao, Yue and Hu, Han and Wei, Yixuan and Zhang, Zheng and Lin, Stephen and Guo, Baining},
  booktitle={Proceedings of the IEEE/CVF international conference on computer vision},
  pages={10012--10022},
  year={2021}
}

@inproceedings{hershey2017cnn,
  title={CNN architectures for large-scale audio classification},
  author={Hershey, Shawn and Chaudhuri, Sourish and Ellis, Daniel PW and Gemmeke, Jort F and Jansen, Aren and Moore, R Channing and Plakal, Manoj and Platt, Devin and Saurous, Rif A and Seybold, Bryan and others},
  booktitle={2017 ieee international conference on acoustics, speech and signal processing (icassp)},
  pages={131--135},
  year={2017},
  organization={IEEE}
}

@inproceedings{gemmeke2017audio,
  title={Audio set: An ontology and human-labeled dataset for audio events},
  author={Gemmeke, Jort F and Ellis, Daniel PW and Freedman, Dylan and Jansen, Aren and Lawrence, Wade and Moore, R Channing and Plakal, Manoj and Ritter, Marvin},
  booktitle={2017 IEEE international conference on acoustics, speech and signal processing (ICASSP)},
  pages={776--780},
  year={2017},
  organization={IEEE}
}

@inproceedings{liu2023audio1,
  title={Audio-visual segmentation by exploring cross-modal mutual semantics},
  author={Liu, Chen and Li, Peike Patrick and Qi, Xingqun and Zhang, Hu and Li, Lincheng and Wang, Dadong and Yu, Xin},
  booktitle={Proceedings of the 31st ACM International Conference on Multimedia},
  pages={7590--7598},
  year={2023}
}

@inproceedings{li2024selm,
  title={SelM: Selective Mechanism based Audio-Visual Segmentation},
  author={Li, Jiaxu and Yu, Songsong and Wang, Yifan and Wang, Lijun and Lu, Huchuan},
  booktitle={Proceedings of the 32nd ACM International Conference on Multimedia},
  pages={3926--3935},
  year={2024}
}

@article{liu2024bavs,
  title={Bavs: Bootstrapping audio-visual segmentation by integrating foundation knowledge},
  author={Liu, Chen and Li, Peike and Zhang, Hu and Li, Lincheng and Huang, Zi and Wang, Dadong and Yu, Xin},
  journal={IEEE Transactions on Multimedia},
  year={2024},
  publisher={IEEE}
}

@inproceedings{li2024qdformer,
  title={Qdformer: towards robust audiovisual segmentation in complex environments with quantization-based semantic decomposition},
  author={Li, Xiang and Wang, Jinglu and Xu, Xiaohao and Peng, Xiulian and Singh, Rita and Lu, Yan and Raj, Bhiksha},
  booktitle={Proceedings of the IEEE/CVF Conference on Computer Vision and Pattern Recognition},
  pages={3402--3413},
  year={2024}
}

@inproceedings{yang2024cooperation,
  title={Cooperation does matter: Exploring multi-order bilateral relations for audio-visual segmentation},
  author={Yang, Qi and Nie, Xing and Li, Tong and Gao, Pengfei and Guo, Ying and Zhen, Cheng and Yan, Pengfei and Xiang, Shiming},
  booktitle={Proceedings of the IEEE/CVF Conference on Computer Vision and Pattern Recognition},
  pages={27134--27143},
  year={2024}
}

@inproceedings{sun2024unveiling,
  title={Unveiling and Mitigating Bias in Audio Visual Segmentation},
  author={Sun, Peiwen and Zhang, Honggang and Hu, Di},
  booktitle={Proceedings of the 32nd ACM International Conference on Multimedia},
  pages={7259--7268},
  year={2024}
}

@inproceedings{selvaraju2017grad,
  title={Grad-cam: Visual explanations from deep networks via gradient-based localization},
  author={Selvaraju, Ramprasaath R and Cogswell, Michael and Das, Abhishek and Vedantam, Ramakrishna and Parikh, Devi and Batra, Dhruv},
  booktitle={Proceedings of the IEEE international conference on computer vision},
  pages={618--626},
  year={2017}
}

@inproceedings{liu2025dynamic,
  title={Dynamic Derivation and Elimination: Audio Visual Segmentation with Enhanced Audio Semantics},
  author={Liu, Chen and Yang, Liying and Li, Peike and Wang, Dadong and Li, Lincheng and Yu, Xin},
  booktitle={Proceedings of the Computer Vision and Pattern Recognition Conference},
  pages={3131--3141},
  year={2025}
}

@article{li2025uni,
  title={Uni-moe: Scaling unified multimodal llms with mixture of experts},
  author={Li, Yunxin and Jiang, Shenyuan and Hu, Baotian and Wang, Longyue and Zhong, Wanqi and Luo, Wenhan and Ma, Lin and Zhang, Min},
  journal={IEEE Transactions on Pattern Analysis and Machine Intelligence},
  year={2025},
  publisher={IEEE}
}

@article{cheng2024mixtures,
  title={Mixtures of experts for audio-visual learning},
  author={Cheng, Ying and Li, Yang and He, Junjie and Feng, Rui},
  journal={Advances in Neural Information Processing Systems},
  volume={37},
  pages={219--243},
  year={2024}
}

@inproceedings{zhudeformable,
  title={Deformable DETR: Deformable Transformers for End-to-End Object Detection},
  author={Zhu, Xizhou and Su, Weijie and Lu, Lewei and Li, Bin and Wang, Xiaogang and Dai, Jifeng},
  year={2021},
  booktitle={International Conference on Learning Representations}
}

@inproceedings{cheng2022masked,
  title={Masked-attention mask transformer for universal image segmentation},
  author={Cheng, Bowen and Misra, Ishan and Schwing, Alexander G and Kirillov, Alexander and Girdhar, Rohit},
  booktitle={Proceedings of the IEEE/CVF conference on computer vision and pattern recognition},
  pages={1290--1299},
  year={2022}
}

@article{zhou2025audio,
  title={Audio-visual segmentation with semantics},
  author={Zhou, Jinxing and Shen, Xuyang and Wang, Jianyuan and Zhang, Jiayi and Sun, Weixuan and Zhang, Jing and Birchfield, Stan and Guo, Dan and Kong, Lingpeng and Wang, Meng and others},
  journal={International Journal of Computer Vision},
  volume={133},
  number={4},
  pages={1644--1664},
  year={2025},
  publisher={Springer}
}

@inproceedings{mao2023multimodal,
  title={Multimodal variational auto-encoder based audio-visual segmentation},
  author={Mao, Yuxin and Zhang, Jing and Xiang, Mochu and Zhong, Yiran and Dai, Yuchao},
  booktitle={Proceedings of the IEEE/CVF International Conference on Computer Vision},
  pages={954--965},
  year={2023}
}

@inproceedings{wang2024prompting,
  title={Prompting segmentation with sound is generalizable audio-visual source localizer},
  author={Wang, Yaoting and Liu, Weisong and Li, Guangyao and Ding, Jian and Hu, Di and Li, Xi},
  booktitle={Proceedings of the AAAI Conference on Artificial Intelligence},
  volume={38},
  pages={5669--5677},
  year={2024}
}

@inproceedings{liu2024audio,
  title={Audio-visual segmentation via unlabeled frame exploitation},
  author={Liu, Jinxiang and Liu, Yikun and Zhang, Fei and Ju, Chen and Zhang, Ya and Wang, Yanfeng},
  booktitle={Proceedings of the IEEE/CVF Conference on Computer Vision and Pattern Recognition},
  pages={26328--26339},
  year={2024}
}

@inproceedings{wang2024can,
  title={Can Textual Semantics Mitigate Sounding Object Segmentation Preference?},
  author={Wang, Yaoting and Sun, Peiwen and Li, Yuanchao and Zhang, Honggang and Hu, Di},
  booktitle={European Conference on Computer Vision},
  pages={340--356},
  year={2024},
  organization={Springer}
}

@article{maaten2008visualizing,
  title={Visualizing data using t-SNE},
  author={Maaten, Laurens van der and Hinton, Geoffrey},
  journal={Journal of machine learning research},
  volume={9},
  number={Nov},
  pages={2579--2605},
  year={2008}
}

@article{mao2025contrastive,
  title={Contrastive conditional latent diffusion for audio-visual segmentation},
  author={Mao, Yuxin and Zhang, Jing and Xiang, Mochu and Lv, Yunqiu and Li, Dong and Zhong, Yiran and Dai, Yuchao},
  journal={IEEE Transactions on Image Processing},
  year={2025},
  publisher={IEEE}
}

@inproceedings{chen2024cpm,
  title={Cpm: Class-conditional prompting machine for audio-visual segmentation},
  author={Chen, Yuanhong and Wang, Chong and Liu, Yuyuan and Wang, Hu and Carneiro, Gustavo},
  booktitle={European Conference on Computer Vision},
  pages={438--456},
  year={2024},
  organization={Springer}
}

@article{riquelme2021scaling,
  title={Scaling vision with sparse mixture of experts},
  author={Riquelme, Carlos and Puigcerver, Joan and Mustafa, Basil and Neumann, Maxim and Jenatton, Rodolphe and Susano Pinto, Andr{\'e} and Keysers, Daniel and Houlsby, Neil},
  journal={Advances in Neural Information Processing Systems},
  volume={34},
  pages={8583--8595},
  year={2021}
}

@inproceedings{du2022glam,
  title={Glam: Efficient scaling of language models with mixture-of-experts},
  author={Du, Nan and Huang, Yanping and Dai, Andrew M and Tong, Simon and Lepikhin, Dmitry and Xu, Yuanzhong and Krikun, Maxim and Zhou, Yanqi and Yu, Adams Wei and Firat, Orhan and others},
  booktitle={International conference on machine learning},
  pages={5547--5569},
  year={2022},
  organization={PMLR}
}

@article{shi2024cross,
  title={Cross-modal cognitive consensus guided audio-visual segmentation},
  author={Shi, Zhaofeng and Wu, Qingbo and Meng, Fanman and Xu, Linfeng and Li, Hongliang},
  journal={IEEE Transactions on Multimedia},
  year={2024},
  publisher={IEEE}
}

@article{dai2022stablemoe,
  title={Stablemoe: Stable routing strategy for mixture of experts},
  author={Dai, Damai and Dong, Li and Ma, Shuming and Zheng, Bo and Sui, Zhifang and Chang, Baobao and Wei, Furu},
  journal={arXiv preprint arXiv:2204.08396},
  year={2022}
}

@article{chen2017deeplab,
  title={Deeplab: Semantic image segmentation with deep convolutional nets, atrous convolution, and fully connected crfs},
  author={Chen, Liang-Chieh and Papandreou, George and Kokkinos, Iasonas and Murphy, Kevin and Yuille, Alan L},
  journal={IEEE transactions on pattern analysis and machine intelligence},
  volume={40},
  number={4},
  pages={834--848},
  year={2017},
  publisher={IEEE}
}

@article{chen2025bootstrapping,
  title={Bootstrapping Audio-Visual Video Segmentation by Strengthening Audio Cues},
  author={Chen, Tianxiang and Tan, Zhentao and Gong, Tao and Chu, Qi and Wu, Yue and Liu, Bin and Yu, Nenghai and Lu, Le and Ye, Jieping},
  journal={IEEE Transactions on Circuits and Systems for Video Technology},
  volume={35},
  number={3},
  pages={2398--2409},
  year={2025},
  publisher={IEEE}
}

@ARTICLE{10979212,
  author={Lv, Ying and Liu, Zhi and Chang, Xiaojun},
  journal={IEEE Transactions on Image Processing}, 
  title={Consistency-Queried Transformer for Audio-Visual Segmentation}, 
  year={2025},
  volume={34},
  number={},
  pages={2616-2627},
  doi={10.1109/TIP.2025.3563076}}

@article{ba2016layer,
  title={Layer normalization},
  author={Ba, Jimmy Lei and Kiros, Jamie Ryan and Hinton, Geoffrey E},
  journal={arXiv preprint arXiv:1607.06450},
  year={2016}
}

@inproceedings{zhou2017scene,
  title={Scene parsing through ade20k dataset},
  author={Zhou, Bolei and Zhao, Hang and Puig, Xavier and Fidler, Sanja and Barriuso, Adela and Torralba, Antonio},
  booktitle={Proceedings of the IEEE conference on computer vision and pattern recognition},
  pages={633--641},
  year={2017}
}

@inproceedings{glisson2025hsda,
  title={HSDA: High-Frequency Shuffle Data Augmentation for Bird's-Eye-View Map Segmentation},
  author={Glisson, Calvin and Chen, Qiuxiao},
  booktitle={2025 IEEE/CVF Winter Conference on Applications of Computer Vision (WACV)},
  pages={8834--8843},
  year={2025},
  organization={IEEE}
}

@article{barbosh2020empirical,
  title={Empirical mode decomposition and its variants: A review with applications in structural health monitoring},
  author={Barbosh, Mohamed and Singh, Premjeet and Sadhu, Ayan},
  journal={Smart Materials and Structures},
  volume={29},
  number={9},
  pages={093001},
  year={2020},
  publisher={IOP Publishing}
}

@article{han2022trusted,
  title={Trusted multi-view classification with dynamic evidential fusion},
  author={Han, Zongbo and Zhang, Changqing and Fu, Huazhu and Zhou, Joey Tianyi},
  journal={IEEE transactions on pattern analysis and machine intelligence},
  volume={45},
  number={2},
  pages={2551--2566},
  year={2022},
  publisher={IEEE}
}

@article{liu2023multi,
  title={Multi-modal mutual attention and iterative interaction for referring image segmentation},
  author={Liu, Chang and Ding, Henghui and Zhang, Yulun and Jiang, Xudong},
  journal={IEEE Transactions on Image Processing},
  volume={32},
  pages={3054--3065},
  year={2023},
  publisher={IEEE}
}

@article{meng2025br,
  title={BR-MoE: Blind Multi-Modal Tracking with Route-Dynamic Mixture of Experts},
  author={Meng, Qingguo and Lu, Andong and Jin, Zhe},
  journal={IEEE Transactions on Circuits and Systems for Video Technology},
  year={2025},
  publisher={IEEE}
}

@inproceedings{diba2023spatio,
  title={Spatio-temporal convolution-attention video network},
  author={Diba, Ali and Sharma, Vivek and Arzani, Mohammad and Van Gool, Luc and others},
  booktitle={Proceedings of the IEEE/CVF International Conference on Computer Vision},
  pages={859--869},
  year={2023}
}

@article{zhang2023fsanet,
  title={FsaNet: Frequency self-attention for semantic segmentation},
  author={Zhang, Fengyu and Panahi, Ashkan and Gao, Guangjun},
  journal={IEEE Transactions on Image Processing},
  volume={32},
  pages={4757--4772},
  year={2023},
  publisher={IEEE}
}

@article{sun2020drcnn,
  title={DRCNN: Dynamic routing convolutional neural network for multi-view 3D object recognition},
  author={Sun, Kai and Zhang, Jiangshe and Liu, Junmin and Yu, Ruixuan and Song, Zengjie},
  journal={IEEE Transactions on Image Processing},
  volume={30},
  pages={868--877},
  year={2020},
  publisher={IEEE}
}

@article{abbas2020biased,
  title={Biased mixtures of experts: Enabling computer vision inference under data transfer limitations},
  author={Abbas, Alhabib and Andreopoulos, Yiannis},
  journal={IEEE Transactions on Image Processing},
  volume={29},
  pages={7656--7667},
  year={2020},
  publisher={IEEE}
}

@article{zhu2025does,
  title={How does audio influence visual attention in omnidirectional videos? Database and model},
  author={Zhu, Yuxin and Duan, Huiyu and Zhang, Kaiwei and Zhu, Yucheng and Zhu, Xilei and Teng, Long and Min, Xiongkuo and Zhai, Guangtao},
  journal={IEEE Transactions on Image Processing},
  year={2025},
  publisher={IEEE}
}

@article{wang2025m2restore,
  title={M2Restore: Mixture-of-experts-based mamba-cnn fusion framework for all-in-one image restoration},
  author={Wang, Yongzhen and Li, Yongjun and Zheng, Zhuoran and Zhang, Xiao-Ping and Wei, Mingqiang},
  journal={IEEE Transactions on Image Processing},
  volume={34},
  pages={8086--8100},
  year={2025},
  publisher={IEEE}
}

@article{zhou2025aloha,
  title={Aloha: Adapting local spatio-temporal context to enhance the audio-visual semantic segmentation},
  author={Zhou, Yang-Hao and Huang, Heyan and Guo, Cunhan and Tu, Rong-Cheng and Xiao, Zeyu and Wang, Bo and Mao, Xian-Ling},
  journal={ACM Transactions on Multimedia Computing, Communications and Applications},
  volume={21},
  number={6},
  pages={1--23},
  year={2025},
  publisher={ACM New York, NY}
}

@article{zhou2025mettle,
  title={Mettle: Meta-token learning for memory-efficient audio-visual adaptation},
  author={Zhou, Jinxing and Li, Zhihui and Yu, Yongqiang and Zhou, Yanghao and Guo, Ruohao and Li, Guangyao and Mao, Yuxin and Han, Mingfei and Chang, Xiaojun and Wang, Meng},
  journal={IEEE Transactions on Pattern Analysis and Machine Intelligence},
  year={2025},
  publisher={IEEE}
}

@inproceedings{huang2025revisiting,
  title={Revisiting audio-visual segmentation with vision-centric transformer},
  author={Huang, Shaofei and Ling, Rui and Hui, Tianrui and Li, Hongyu and Zhou, Xu and Zhang, Shifeng and Liu, Si and Hong, Richang and Wang, Meng},
  booktitle={2025 IEEE/CVF Conference on Computer Vision and Pattern Recognition (CVPR)},
  pages={8352--8361},
  year={2025},
  organization={IEEE}
}

@inproceedings{du2025crab,
  title={Crab: A unified audio-visual scene understanding model with explicit cooperation},
  author={Du, Henghui and Li, Guangyao and Zhou, Chang and Zhang, Chunjie and Zhao, Alan and Hu, Di},
  booktitle={2025 IEEE/CVF Conference on Computer Vision and Pattern Recognition (CVPR)},
  pages={18804--18814},
  year={2025},
  organization={IEEE}
}

@article{zhang2022c,
  title={{C$^{2}$DFNet}: Criss-cross dynamic filter network for {RGB-D} salient object detection},
  author={Zhang, Miao and Yao, Shunyu and Hu, Beiqi and Piao, Yongri and Ji, Wei},
  journal={IEEE Transactions on Multimedia},
  volume={25},
  pages={5142--5154},
  year={2022},
  publisher={IEEE}
}

@article{piao2022semi,
  title={Semi-supervised video salient object detection based on uncertainty-guided pseudo labels},
  author={Piao, Yongri and Lu, Chenyang and Zhang, Miao and Lu, Huchuan},
  journal={Advances in neural information processing systems},
  volume={35},
  pages={5614--5627},
  year={2022}
}

@inproceedings{ren2022adaptive,
  title={Adaptive co-teaching for unsupervised monocular depth estimation},
  author={Ren, Weisong and Wang, Lijun and Piao, Yongri and Zhang, Miao and Lu, Huchuan and Liu, Ting},
  booktitle={European Conference on Computer Vision},
  pages={89--105},
  year={2022},
  organization={Springer}
}

@article{zhang2022exploring,
  title={Exploring spatial correlation for light field saliency detection: Expansion from a single view},
  author={Zhang, Miao and Xu, Shuang and Piao, Yongri and Lu, Huchuan},
  journal={IEEE Transactions on Image Processing},
  volume={31},
  pages={6152--6163},
  year={2022},
  publisher={IEEE}
}

\begin{IEEEbiography}[{\includegraphics[width=25mm,height=35mm,clip,keepaspectratio]{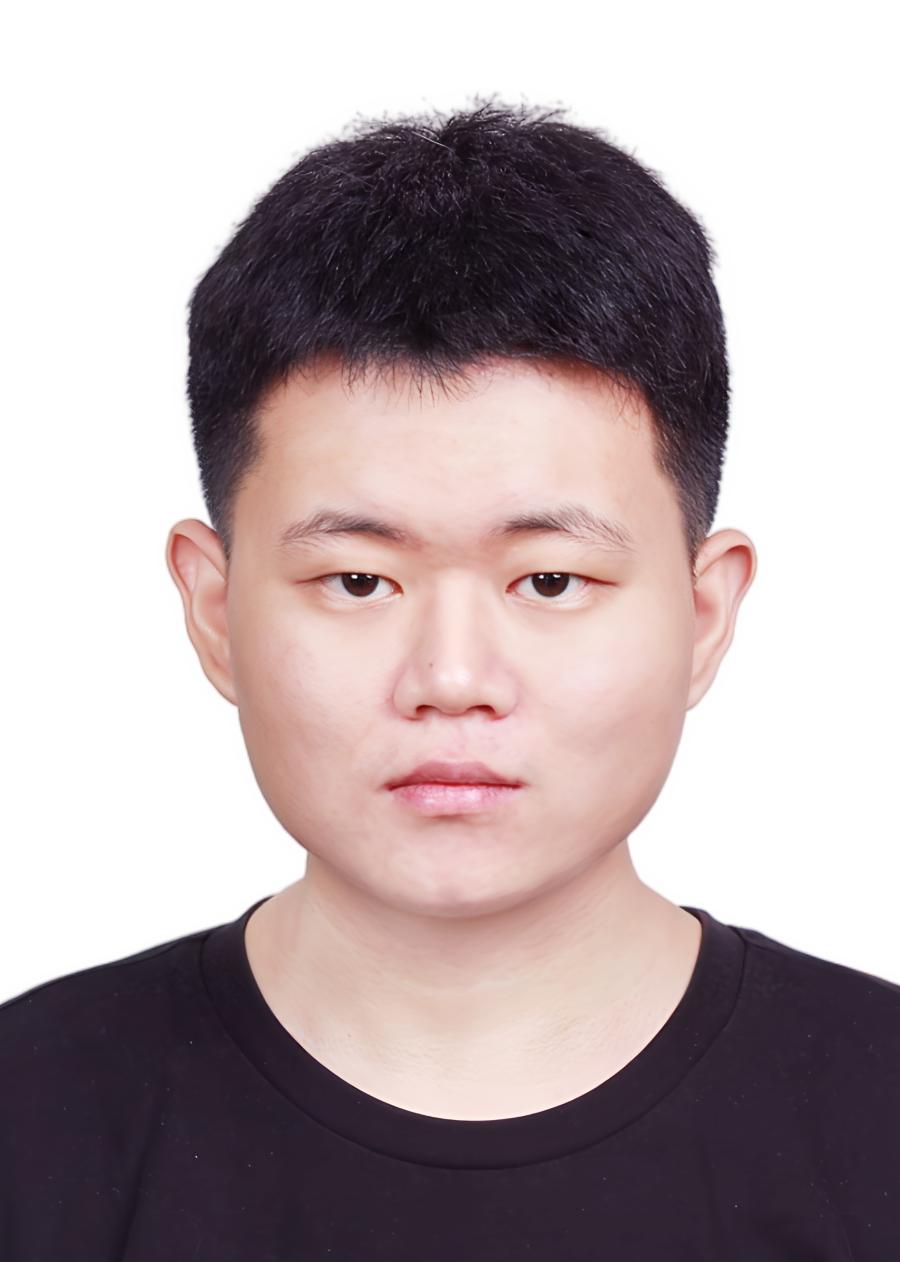}}]{Kai Peng}
received the B.S. degree in computer science and technology from Xinjiang University in 2024.

He is currently a Ph.D. student with the School of Information and Communication Engineering, Dalian University of Technology, Dalian, China. His research interests include computer vision, audio-visual segmentation, and multimodal learning.
\end{IEEEbiography}
\begin{IEEEbiography}[{\includegraphics[width=25mm,height=35mm,clip,keepaspectratio]{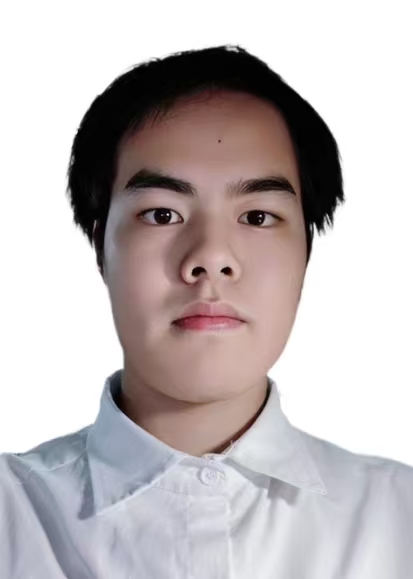}}]{Yunzhe Shen}
	received the B.S. degree in software engineering from Dalian University of Technology, Dalian, China, in 2024. 

    He is currently with IIAU-OIP Lab, Dalian University of Technology, Dalian, China. His current research interests include computer vision and multimodal learning.
\end{IEEEbiography}
\begin{IEEEbiography}[{\includegraphics[width=25mm,height=35mm,clip,keepaspectratio]{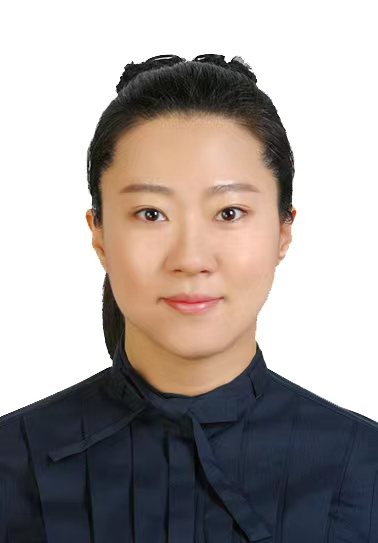}}]{Miao Zhang}
	(Member, IEEE) received the B.S. degree in computer science from the Memorial University of Newfoundland, St. John’s, NL, Canada, in 2005, and the Ph.D. degree in electronic engineering from Kwangwoon University, Seoul, South Korea, in 2012. From 2013 to 2015, she was an Assistant Professor with the Department of Game and Mobile Contents, Keimyung University, Daegu, South Korea, and an Adjunct Professor with the Department of Computer Science, DigiPen Institute of Technology, Redmond, WA, USA, respectively.
	
	She is currently an Associate Professor with the Key Laboratory for Ubiquitous Network and Service Software of Liaoning Province, DUT-RU International School of Information Science and Software Engineering, Dalian University of Technology, Dalian, China. Her research interests include computer vision, machine learning, and 3D imaging and visualization.
\end{IEEEbiography}
\begin{IEEEbiography}[{\includegraphics[width=25mm,height=35mm,clip,keepaspectratio]{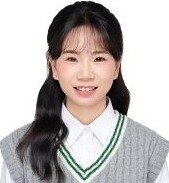}}]{Jia Wang}
received the B.E. degree in software engineering from Chongqing
Normal University, Chongqing, China, in 2020, and the M.S. degree
in computer technology from Chongqing University of Posts and
Telecommunications, Chongqing, China, in 2023.

She is currently a Ph.D. student in software engineering at
Dalian University of Technology, Dalian, China. Her research
interests include super-resolution and medical image analysis.
\end{IEEEbiography}

\begin{IEEEbiography}[{\includegraphics[width=25mm,height=35mm,clip,keepaspectratio]{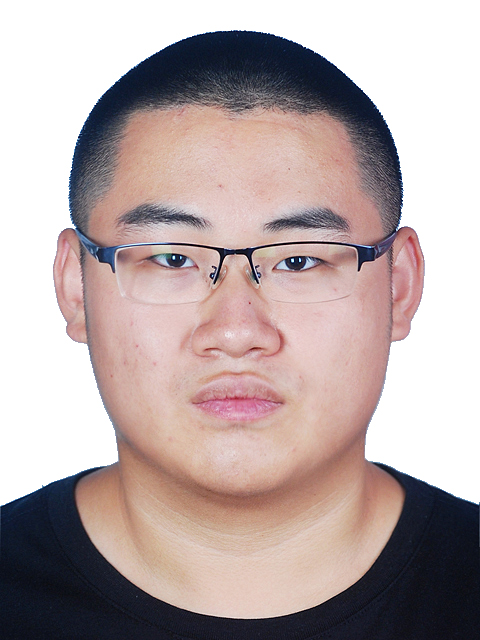}}]{Leiye Liu} 
    received the bachelor's degree from Dalian University of Technology. 
    
    He is currently a Ph.D. Student at Dalian University of Technology. His research interests include segmentation, state space model, foundation model, and multimodal perception.
\end{IEEEbiography}

\begin{IEEEbiography}[{\includegraphics[width=25mm,height=35mm,clip,keepaspectratio]{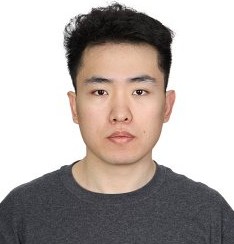}}]{Wei Ji}
    received the Ph.D. degree from the University of Alberta. He was a Visiting Ph.D. student at Johns Hopkins University. He is currently a Postdoctoral Researcher with Yale University. 
    
    His research interests include segmentation, visual saliency, foundation model, multimodal perception, and medical image analysis. He achieved the CVPR Best Paper Candidate and the MICCAI Young Scientist Award Nominee.
\end{IEEEbiography}
\begin{IEEEbiography}[{\includegraphics[width=25mm,height=35mm,clip,keepaspectratio]{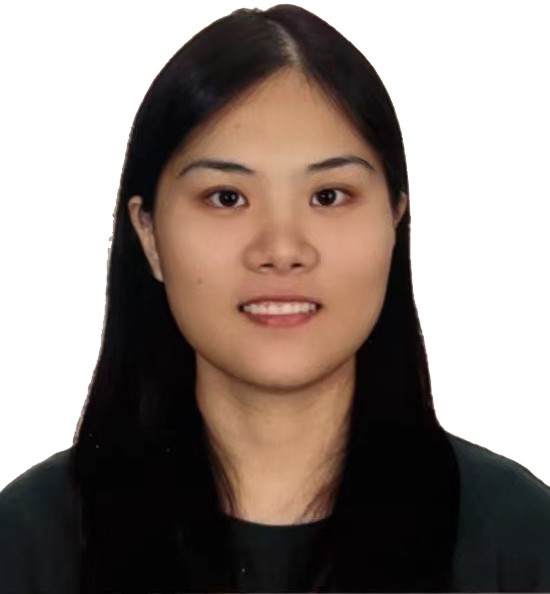}}]{Jingjing Li}
    is currently a Postdoctoral Research Associate at Carnegie Mellon University, USA. She received her Ph.D. degree from the University of Alberta, Canada. Her research interests include computer vision, multimodal large language models, scene understanding, and medical image analysis.
\end{IEEEbiography}
\begin{IEEEbiography}[{\includegraphics[width=25mm,height=35mm,clip,keepaspectratio]{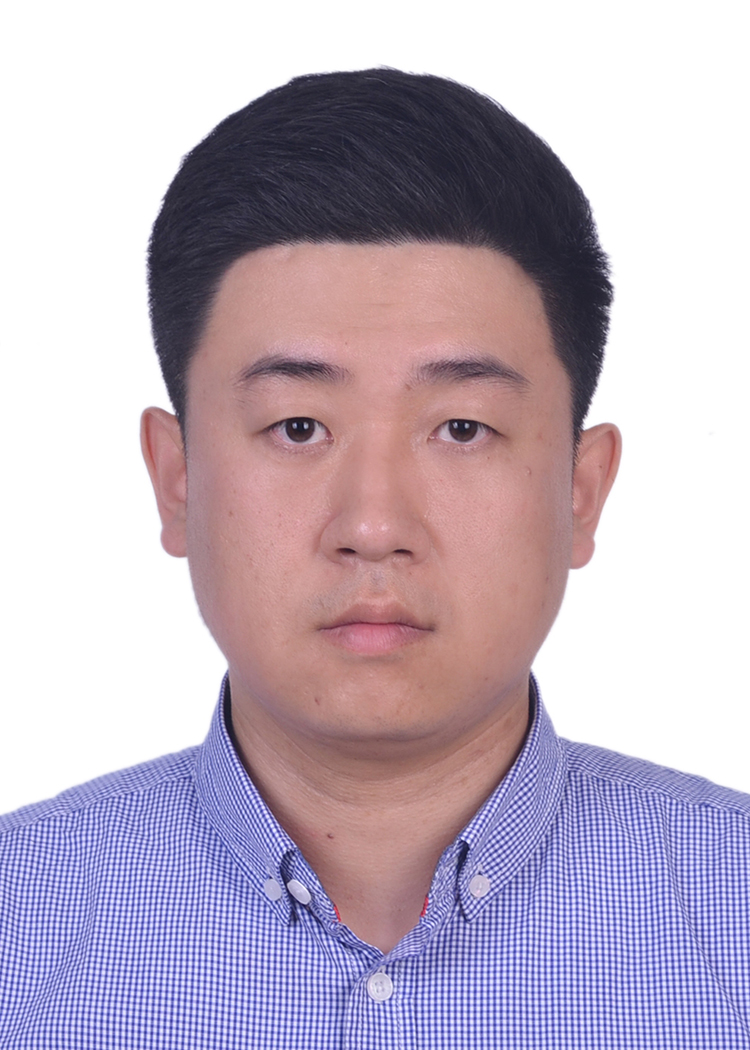}}]{Yongri Piao}
	(Member, IEEE) received the M.Sc. and Ph.D. degrees in information and communication engineering from Pukyong National University, Busan, South Korea, in 2005 and 2008, respectively. 
	
	Since 2012, he has been an Associate Professor of information and communication engineering with the Dalian University of Technology, Dalian, China. His research interests include 3D computer vision and sensing, object detection and target recognition, and 3D reconstruction and visualization.
\end{IEEEbiography}
\begin{IEEEbiography}[{\includegraphics[width=25mm,height=35mm,clip,keepaspectratio]{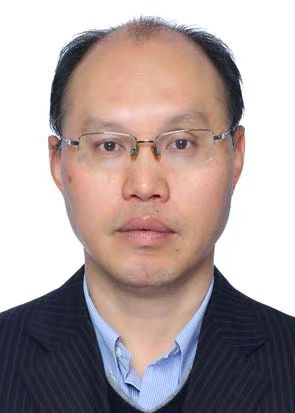}}]{Huchuan Lu}
	(Fellow, IEEE) received the M.S. degree in signal and information processing and the Ph.D. degree in system engineering from the Dalian University of Technology (DUT), Dalian, China, in 1998 and 2008, respectively. 
	
	He joined the Faculty of the School of Information and Communication Engineering, DUT, in 1998, where he is currently a Full Professor. His research interests include computer vision and pattern recognition with a focus on visual tracking, saliency detection, and segmentation.
\end{IEEEbiography}

\vfill
\end{document}